\documentclass[10pt,twocolumn,letterpaper]{article}

\usepackage{cvpr}
\usepackage{times}
\usepackage{epsfig}
\usepackage{graphicx}
\usepackage{subfigure}
\usepackage{float}
\usepackage{amsmath}
\usepackage{amssymb}
\usepackage{appendix}

\usepackage[breaklinks=true,bookmarks=false]{hyperref}

\cvprfinalcopy 

\def\cvprPaperID{****} 

\begin{document}

\title{Deeply Aligned Adaptation for Cross-domain Object Detection}

\author{Minghao Fu,  Zhenshan Xie,  Wen Li,  Lixin Duan\\
University of Electronic Science and Technology of China\\
{\tt\small \{minhfu2,ivanxie1022,liwenbnu,lxduan\}@gmail.com}
}

\maketitle

\begin{abstract}
Cross-domain object detection has recently attracted more and more attention for real-world applications, since it helps build robust detectors adapting well to new environments. In this work, we propose an end-to-end solution based on Faster R-CNN, where ground-truth annotations are available for source images (\textit{e.g.}., cartoon) but not for target ones (\textit{e.g.}, watercolor) during training. Motivated by the observation that the transferabilities of different neural network layers differ from each other, we propose to apply a number of domain alignment strategies to different layers of Faster R-CNN, where the alignment strength is gradually reduced from low to higher layers. Moreover, after obtaining region proposals in our network, we develop a foreground-background aware alignment module to further reduce the domain mismatch by separately aligning features of the foreground and background regions from the source and target domains. Extensive experiments on benchmark datasets demonstrate the effectiveness of our proposed approach.
\end{abstract}

\section{Introduction}

Object detection, as one of the fundamental problems in computer vision, aims to classify objects in an input image and localize them with bounding boxes~\cite{liu1809deep}. With the advance of deep convolutional neural network~\cite{he2016deep,lecun2015deep,simonyan2014very}, deep learning based models become the mainstream solutions for object detection, and many detection models have been proposed in recent years~\cite{girshick2014rich,girshick2015fast,ren2015faster,redmon2016you,liu2016ssd,lin2017feature,zhou2019objects}.

Although a great progress has been made, training deep object detection models relies on 
a large number of labeled training data. Moreover, it has also been observed that the performance often drops significantly when the learnt models being applied to a new scenario. To alleviate the burden of labeling a large volume training data for every new scenario, there is recently an increasing interest in developing effective approaches to adapt the detectors from a label-rich source domain to a label-scarce target domain, which is also known as \textit{cross-domain object detection}~\cite{chen2018domain,inoue2018cross,he2019multi,saito2019strong,Cai_2019_CVPR,xu2019wasserstein,Kim_2019_CVPR,Zhu_2019_CVPR,SHAN201931}. 

However, learning a model for cross-domain object detection is a nontrivial task. Detection models are vulnerable to data variance,  due to simultaneously predicting the bounding box and object class. A small variance in an image may lead to inaccurate bounding box and wrong classification~\cite{wang2018deep}. Furthermore, in real-world tasks, the application scenarios may differ in many aspects, for example, object appearance, complex background, environment illumination, \textit{etc}. As a result, straightforwardly applying traditional domain adaptation strategies can hardly produce satisfactory performance. It is more desirable to align the data distributions in fine levels for handling the diverse domain shift.

To this end, in this paper, we propose a Deeply Aligned Adaptation (DAA) approach for cross-domain object detection. Our work is partially inspired by~\cite{yosinski2014transferable}, where they observed that the transferability of features learned at different layers varies, and features from bottom layers are more general while features from top layers are more specific. Therefore, we propose to apply different domain alignment strategies for different layers, and \emph{gradually} reduce alignment strength from low-level layers to high-level layers. As a good practice, we employ adversarial training at each layer to align the source and target distributions. The adversarial training part is performed on local patch features at bottom layers, while on global image feature at top layers. To cope with the transition of transferability from bottom to top layers, we also introduce a novel transition module for middle layers to bridge the local and global alignment in a smooth way.  

Moreover, to further boost the generalization ability of local features at bottom layers, we introduce a domain mask integration (DMI) module for enhancing the adversarial training process. The DMI module automatically assigns high weights for samples more useful for optimizing the discriminator and generator, resulting in a finer distribution alignment and more general features. On the other hand, the normal adversarial training is used for top layers for a relatively weaker alignment. And the transition module keeps both types of alignments for middle layers.

Additionally, to handle the diverse domain shift across different detection scenarios, we further introduce a new foreground-background aware alignment for improving the distribution alignment on instance features. In particular, features extracted based on region proposals often contain a considerable portion of background regions. Due to the large diversity in instance appearances across domains, directly performing distribution alignment on ROI features between two domains often mistakenly aligns foreground regions in one domain to background regions in the other domain. Thus, we propose to perform feature alignment for foreground and background proposals separately to avoid the aforementioned misalignment between them. Overall, the main contributions of this paper can be summarized as follows:
\vspace{-7pt}
\begin{itemize}
 \item We propose a new transition module for middle layers to bridge the local patch feature alignment and the global feature alignment, which also copes with the the transition of transferability from bottom to top layers. 

 \item  For local patch alignment, we introduce a domain-mask integration module to enhance the optimization of both discriminator and generator for producing a more general feature representation. 

 \item For instance alignment, we further propose a foreground-background aware alignment module to alleviate the misalignment between foreground and background regions from the source and target domains. 

 \item We implement the above modules into an end-to-end domain adaptive object detection model based on Faster R-CNN. With extensive experiments conducted on multiple benchmark datasets, we validate the effectiveness of our newly proposed model which generally outperforms the existing state-of-the-art methods.
\end{itemize}

\section{Related Work}
\noindent{\bf Domain Adaptation:}  The goal of domain adaptation is to adapt information learned from the source domain to the target domain of interest, where there are very limited or even no labeled target samples available for training.
Till now, quite a few methods~\cite{ganin2014unsupervised,saito2018maximum,long2016unsupervised,saenko2010adapting,tzeng2017adversarial,tzeng2014deep} have been proposed to match feature distributions between the source and target domains to reduce the domain discrepancy in visual applications as image classification, object detection and image segmentation.

With the advances of adversarial learning~\cite{goodfellow2014generative,tzeng2017adversarial}, the domain confusion for feature alignment can also be done by adversarial learning based on feature representations. Motivated by that, in this work we propose to regularize the domain classifier by considering the transferability difference among network layers, where the corresponding feature alignments are enforced with different strength.

\noindent{\bf Object Detection:} As an important research direction, object detection has been widely studied for years. Conventional methods~\cite{felzenszwalb2009object,dalal2005histograms} use sliding windows for instance recognition and localization. With the fast progress of deep learning techniques, various methods have been proposed, and the performance of object detection has reached an unprecedented high level~\cite{girshick2014rich,girshick2015fast,ren2015faster,redmon2016you,liu2016ssd,lin2017feature,zhou2019objects}. Those methods can be generally divided into two categories: one-stage and two-stage. 

As one of the most representative two-stage method, the effectiveness and efficiency of Faster R-CNN~\cite{ren2015faster} have been demonstrated. It basically generates coarse proposals with Region Proposal Networks (RPN) at the first stage and feeds the proposals into a refinement module at the second stage. Due to its good performance and well organized network structure, we adopt Faster R-CNN as the backbone in our proposed method.

\noindent{\bf Cross-domain Object Detection:} The problem of domain shift in object detection is much more challenging than that in image classification, due to complex backgrounds, existence of multiple objects, \textit{etc}. So far, this topic has not been studied much in the literature, and researchers have proposed a few approaches to alleviate the cross-domain issue. DA-Faster~\cite{chen2018domain} firstly tackled the problem by dividing the domain shift into image-level and instance-level cases, for which two domain classifiers are applied to perform domain adversarial learning respectively. Shan \textit{et al.}~\cite{SHAN201931} explored to use Generative Adversarial Network (GAN)~\cite{goodfellow2014generative} for pixel-level adaptation. Besides, DM~\cite{Kim_2019_CVPR} proposed to deal with both pixel-level translation and source-biased feature discriminability simultaneously.  Moreover, in~\cite{inoue2018cross}, a weakly-supervised approach was proposed to progressively transfer a model trained only in the source domain to the target domain, where the first step is to train with images rendered by CycleGAN~\cite{zhu2017unpaired} and then to perform adaptation by using pseudo labels. And SWDA~\cite{saito2019strong} took strong and weak alignment strategies for local and global features, respectively. In this work, we follow the same unsupervised domain adaptation setting as in the above methods. We propose different alignment strategies for different transferabilities of the network layers, and our DAA achieves state-of-the-art performance.

\section{Proposed Method}

 We consider the unsupervised domain adaptation setting. Specifically, the training data consists of two domains, a source domain which contains images annotated with bounding boxes on instances therein, and a target domain which contains only images without any annotations. We aim to train an object detector which performs well in the target domain, though the relevant bounding boxes are unavailable in the target domain training data. 
 
 Formally, we denote the source domain by $S = \{x_i^s|^{n_s}_{i=1}\}$, where $n_s$ is the number of source images, and each $x_i^s$ is a source training sample which is annotated with a set of bounding boxes and corresponding class labels. We also denote the bounding boxes of image $x_i^s$ by  $b_i^s \in \mathbb{R}^{k \times 4}$, where $k$ is the number of boxes and each box is represented by 4-dimensional coordinates. Similarly, the set of class labels is denoted by $c_i^s$, where  $c_i^s \in \mathbb{R}^{k \times 1}$ with each element being the class label for one box. The target domain contains only images without annotation, which can be represented as  $T = \{x_j^t|_{j=1}^{n_t}\}$, where $n_t$ is the number of target samples.
 
\subsection{Network Architecture Overview}
As discussed in the Introduction, detection models are vulnerable to data variance, due to simultaneously predicting the bounding box and object class. On the other hand, the real domain shift occurs on many aspects, \emph{e.g.}, object appearance, complex background, environment illumination, etc. 

\begin{figure*}[ht]
\centering
\includegraphics[scale=0.4]{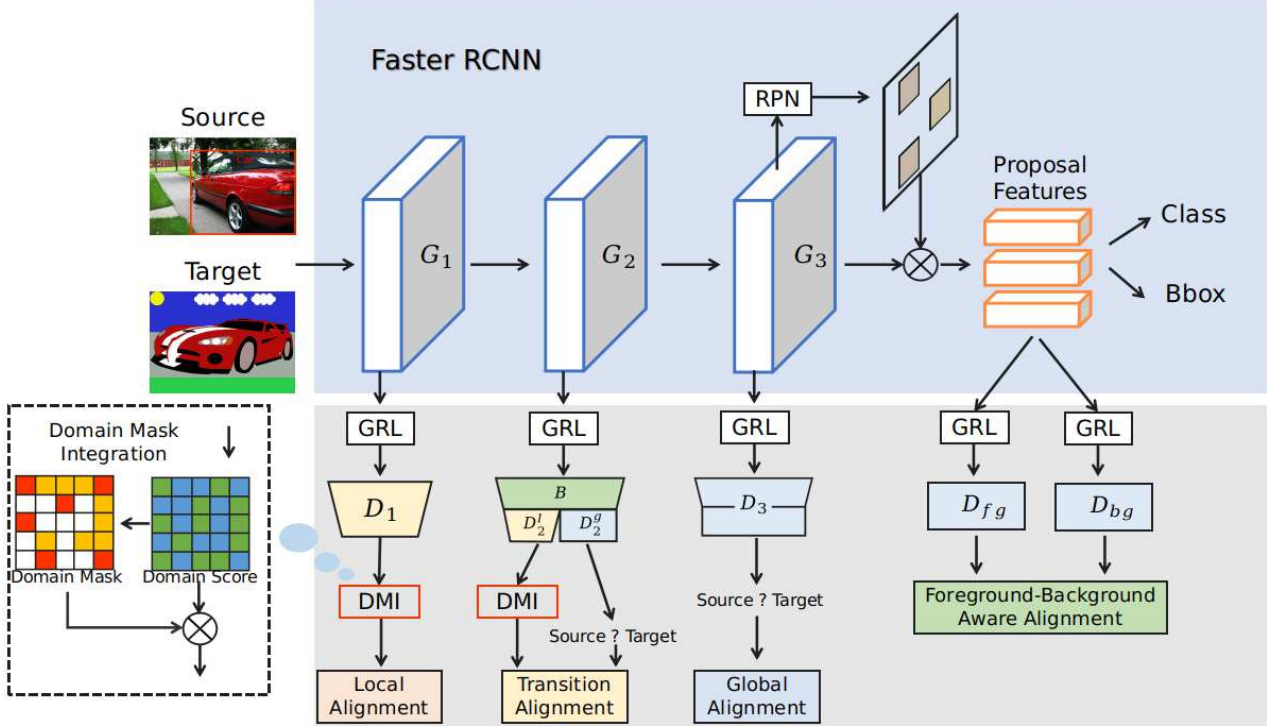}
\caption{\small Network structure of Deeply Aligned Adaptation (DAA) model. Our network is based on Faster R-CNN, in which $G_1$, $G_2$, $G_3$ are three convolutional blocks. We perform domain adaptation on both feature level and proposal level. For feature level,  three modules are introduced, \textit{local}, \textit{transition} and \textit{global} adaptor, respectively. The \textit{domain mask integration (DMI)} module is further deployed to local and transition adaptor for enhancing patch-level feature alignment. For the proposal level, we  employ the \textit{foreground-background aware alignment} to individually adapt  foreground and background proposals for a more robust distribution alignment.}
\label{fig:1}
\end{figure*}

To address these challenges, we propose a Deeply Aligned Adaptation (DAA) model for cross-domain object detection. The architecture of our proposed model is shown in  Fig.~\ref{fig:1}. We utilize Faster R-CNN~\cite{ren2015faster} as the basic object detector. Inspired by \cite{chen2018domain}, we also perform alignment on both image level and instance level, however, with substantial improvements for strengthening the domain alignments. In particular, to cope with the gradually shifting feature transferability of CNNs, we propose a \textbf{{transition feature  distribution alignment}} module and a \textbf{{domain mask integration}} module for image level alignment. For instance level alignment, we further design a \textbf{{foreground-background aware alignment}} module to address the potential issue of foreground-background misalignment caused by the large diversity in object appearances, poses, \textit{etc}. In the following, we will simply review the Faster R-CNN approach, and then introduce our three new modules followed by a summarization of the whole model. 

\noindent\textbf{Faster R-CNN:} As shown in the blue region of Fig.~\ref{fig:1}, the Faster R-CNN model contains a feature extractor, a region proposal network (RPN), and a prediction head. We denote the backbone feature extractor by $G$, which usually is a CNN networks containing convolutional layers and activation layers (\emph{e.g.}, VGG~\cite{simonyan2014very} or ResNet~\cite{He_2016_CVPR}). In Fig.~\ref{fig:1}, we decompose the feature extractor $G$ into three blocks $G = \{G_1, G_2, G_3\}$ for convenience of presenting our new modules later, and use $G$ to denote all three blocks. 
The feature extractor takes an image $x$ as input, and produces an activation map $G(x)$. Then the subsequent RPN network $R$ takes $G(x)$ as input to generate a set of object proposals. These proposals will be used to pool the corresponding regions of activation map into instance-level features with fixed feature dimension, \emph{i.e.}, ROI-pooling. We use $R(G(x))$ to represent all instance-level features after ROI-pooling for simplicity. Finally, the prediction head takes the instance-level features as input, and predicts their bounding boxes and class labels. The training loss for the prediction head can be written as:

\vspace{-10pt}
\begin{equation}
\begin{split}
L_{det} = \frac{1}{n_s}\sum_{i=1}^{n_s}\Big(&L_{cls}\big(R(G(x_i^s)), c_i^s\big) + \\ &L_{bbox}\big(R(G(x_i^s)),b_i^s\big)\Big) 
\end{split}
\end{equation}
where $L_{cls}$ and $L_{bbox}$ denote the classification and localization loss, respectively.  Note that we omit the loss for RPN in the discussion above and afterwards for simplicity, but it is also involved in the training phase as same as the original Faster R-CNN model.

\subsection{Transitional feature distribution alignment}
As indicated in \cite{yosinski2014transferable}, the transferabilities of features learned at different network layers vary, and the features from bottom layers are often local and more general, while features from higher layers are relatively more global and specific. In the context of cross-domain object detection, the recent work~\cite{saito2019strong} made attempt to use different losses in adversarial training for bottom and top layers to model strong and weak alignments, respectively. However, the middle layers were not considered, which are actually important for an effective cross-domain distribution alignment, since the feature transferability shifts down \emph{gradually} when the layer goes deeper in CNNs.

In this work, we propose a new  method for feature-level alignment by jointly using local, transition and global adaptors. The local adaptor is used on bottom layers for aligning the distributions of local patch features, while the global adaptor is used on the top layers for aligning the global image features as a whole. Different from~\cite{saito2019strong}, which relaxes the global alignment on top layers, we go oppositely to strengthen the alignment of local features with a new domain mask integration module (see Section~\ref{DMI} for details), thus leading to a more effective distribution alignment. To be inline with the gradual shift of feature transferability, we additionally propose a transition alignment module for middle layers to bridge the local and global alignments.

\noindent\textbf{Local Adaptor:} The local adaptor is deployed at the bottom block $G_1$ to align the distribution of the activation between two domains. Since each activation after the bottom block corresponds to a local image patch, it can be deemed as to align the distributions of local image patches. As a good practice, we use the adversarial training strategy for the distribution alignment. 

Specifically, in adversarial training, we use a discriminator $D$ to classify each sample into the source domain or the target domain. Meanwhile, the feature extractor $G$ is trained to confuse the discriminator with adversarial training. Intuitively, if samples cannot be easily distinguished, the feature representation from $G$ is more likely to be domain invariant, and thus generalizes well to the target domain. Formally, let us denote $d \in \{0, 1\}$ as the domain label where $0$ is the source domain and $1$ is the target domain, then the loss for adversarial training can be written as:
\begin{equation}
\small
\label{adv}
L_{adv}  = \sum_{x} -d\log(D(G(x))) - (1-d) \log (1-D(G(x)))        
\end{equation}

Using adversarial training means we need to train $D$ to minimize the above loss, and at the same time train $G$ to maximize it. As shown in~\cite{ganin2014unsupervised}, this can be simply achieved by inserting a Gradient Reverse Layer (GRL) between $G$ and $D$ which reverses the sign of gradients during back-propagation. Then, one can minimize the above loss as usual to perform adversarial training. 

For our local adaptor, the distribution alignment is performed on local patch features. As shown in Fig~\ref{fig:1}, the discriminator $D_1$ consists of several $1\times 1$ convolutional layers. It takes the feature map from $G_1$ as input, and outputs a domain probability map accordingly. 
Given an image $x_i$, let us denote $\mathbf{Z}_i^{(1)} = G_1(x_i)$ as the feature map after $G_1$, then the domain probability map generated by $D_1$ can be denoted as $D_1(\mathbf{Z}_i^{(1)})$. We also use  $D_1(\mathbf{Z}_i^{(1)})_{w, h}$ to represent the domain probability score at position $(w, h)$ of the domain probability map, where $1\leq w \leq W_1$ and $1\leq h \leq H_1$ with $W_1$ and $H_1$ being the width and height of the domain probability map. 

The training objective of the local adaptor can be written as:
\vspace{-10pt}
\begin{align}
L_{loc}(x_i) = \frac{1}{H_1 W_1} \sum_{h=1}^{H_1} \sum_{w=1}^{W_1} L_{ce}\big(D_1(\mathbf{Z}_{i}^{(1)})_{w,h}, d_i\big)
\end{align}
where  $L_{ce}$ is the cross entropy loss similarly defined as in (Eq.~(\ref{adv})). Since we expect local patch features to be more general, as shown in Fig.~\ref{fig:1}, we further design a domain mask integration module to enhance the distribution alignment, which will be explained in detail in Section~\ref{DMI}.

\noindent\textbf{Global Adaptor:}
For top layers, we apply the global alignment to match the distribution of two domains on image-level features.  Different from the local adaptor, the discriminator $D_3$ outputs a single value for each input image.  Similar to the local alignment, we denote $\mathbf{Z}^{(3)}_i = G_3(G_2(G_1(x_i)))$ as the feature map of $x_i$ produced by $G_3$. Then, the loss for global alignment can be written as:
\begin{align}
L_{global}(x_i) = L_{ce}\big(D_3(\mathbf{Z}_i^{(3)}), d_i\big)
\end{align}

\noindent\textbf{Transition Adaptor:}
To bridge the local and global alignments, we introduce a transition adaptor for the middle block $G_2$ to smooth the transformation process of CNN. As shown in Fig.~\ref{fig:1}, we use a Y-shape architecture for the discriminator $D_2$ to integrate the local and global alignments. The feature map generated by $G_2$ will be passed to a shared convolutional block $B$, and then go through several convolutional layers to generate a domain probability map for local alignment, or several fully connected layers to output a single domain probability for global alignment, respectively. 

Similarly as above, we denote the feature map of $x_i$ after $G_2$ as $\mathbf{Z}_i^{(2)} = G_2(G_1(x_i))$, and use $D_2^{l}(B(\mathbf{Z}_i^{(2)}))$ and $D_2^{g}(B(\mathbf{Z}_i^{(2)}))$ to represent the outputs from local and global alignment branches, respectively. We also denote $D_2^{l}(B(\mathbf{Z}_i^{(2)}))_{w, h}$  as the domain probability score at position $(w, h)$ of the domain probability map from the local alignment branch. 

Then, the loss function of transition adaptor is written as:
\begin{equation}
\begin{split}
L_{tr}(x_i)  =  \frac{1}{H_2 W_2} \sum_{h=1}^{H_2} \sum_{w=1}^{W_2}  &L_{ce}\big(D_2^{l}(B(\mathbf{Z}^{(2)}_{i}))_{w, h}, d_{i}\big ) \\
+ &L_{ce}\left(D_2^{g}\left(B(\mathbf{Z}^{(2)}_{i})\right), d_i\right)
\end{split}
\end{equation}
where $W_2$ and $H_2$ are the width and height of domain probability map $D_2^{l}(B(\mathbf{Z}_i^{(2)}))$. By integrating local and global feature distributions alignment in one module, the domain classifier could simultaneously extract information in both local and global views then fuse corresponding features to express middle-level semantics and have the effect of feature alignment in intermediate feature space. 

\noindent\textbf{Summary:} The overall loss function of transitional feature distribution alignment is summarized as:
\begin{align}
L_{feat}(x_i) = L_{loc}(x_i) + L_{tr}(x_i) + L_{global}(x_i)
\end{align}

\subsection{Domain Mask Integration for Local Alignment} \label{DMI}
Local features are expected to be more general, while global features are more specific. Consequently, it is desirable to impose a relative stronger distribution alignment for local features, while a weaker one for global features in the domain adaptation scenario. The previous work SWDA~\cite{saito2019strong} implemented this by relaxing the global domain alignment with a focal loss. However, this also possibly reduces the effect of adversarial training, leading to inferior domain distribution alignment. To this end, we keep the adversarial training for global feature unchanged, and instead  design a new domain mask integration module to strengthen the local feature alignment. 

Our motivation is from the observation on the optimization process with GRL. Specifically, an ideal convergence point of the domain adversarial training is that most training samples cannot be easily distinguished, which means the predictions from discriminator on these samples are around the middle point $0.5$. We are specifically interested in the samples being predicted far away from the middle point. 

\begin{figure}[t]
\centering
\includegraphics[height=4.5cm]{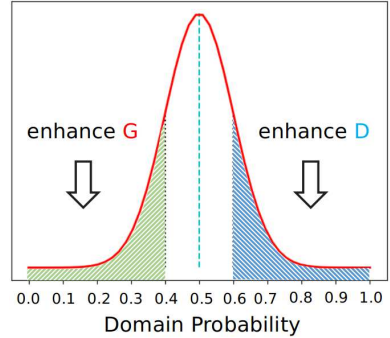}
\caption{\small Illustration of using domain probability to enhance adversarial training in our domain mask integration. The red curve represents the distribution of domain probabilities of source samples. Samples in the left part are mostly correctly classified by the discriminator, which are useful for enhancing feature extractor $G$, while the samples in the right part are mostly mis-classified, which are useful for the discriminator $D$.}
\vspace{-10pt}
\label{fig:dmi_example}
\end{figure}

Taking the source sample $x^s$ as an example, if the prediction is close to $1$, it means the discriminator predicts $x^s$ more likely to be a target sample, which will lead to a high cross entropy loss according to (Eq. ~(\ref{adv})). During the training process, such samples would be more important to the discriminator than others, because they can effectively strengthen the discriminator via back-propagation. On the other hand, when the prediction is close to $0$, it means the discriminator does a good job, and correctly predicts $x^s$ as a source sample. In this case, the sample does not affect the discriminator too much, as it produces a low cross-entropy loss. However, the sample would be very useful for the feature extractor $G$, since in back-propagation, the gradient generated by this sample will be reversed by the GRL to guide $G$ to generate domain-invariant features. We illustrate our analysis in Fig~\ref{fig:dmi_example}. The above analysis can be applied to the target samples similarly. 

Therefore, we propose to reweight training samples (i.e., local patch features) to give higher weights for samples with low or high prediction scores by the discriminator. In this way, we expect to boost both the discriminator and the feature extractor, thus aligning the distributions of two domains more precisely through adversarial training. In particular, given a sample $x$ and its domain probability score predicted by domain discriminator $p = D(G(x))$, we then calculate its weight as follows:
\begin{align}
m = \eta \cdot |p - 0.5| + 1
\label{eqn:domain_score}
\end{align}
where $\eta$ is a hyper-parameter standing the scale of domain mask and $|.|$ denotes the $l_1$ distance. 

Given an image $x_i$, we compute the weights for all patch features in the local alignment module as defined in (Eq.~(\ref{eqn:domain_score})), and then obtain a weight map $M^{(1)}$ with size $W_1\times H_1$, which we refer to as a \emph{domain probability mask}. We also denote $M^{(1)}_{w, h}$ as the weight for patch feature at position $(w, h)$. The loss for local alignment can be updated by integrating the domain probability mask as follows:
\begin{equation}
\small
L_{loc}(x_i) = \frac{1}{H_1 W_1} \sum_{h=1}^{H_1} \sum_{w=1}^{W_1}  M^{(1)}_{(w,h)} L_{ce}\big(D_1(\mathbf{Z}_{i}^{(1)})_{w,h}, d_i\big) 
\end{equation}
Similarly, we also integrate the domain probability mask into the local alignment branch in the transition adaptor as illustrated in Fig.~\ref{fig:1}.

\subsection{Foreground-background Aware Alignment}
\begin{figure}[t]
\centering
\subfigure[Before Adaptation]{
\label{fb_before}
\includegraphics[width=.31\linewidth]{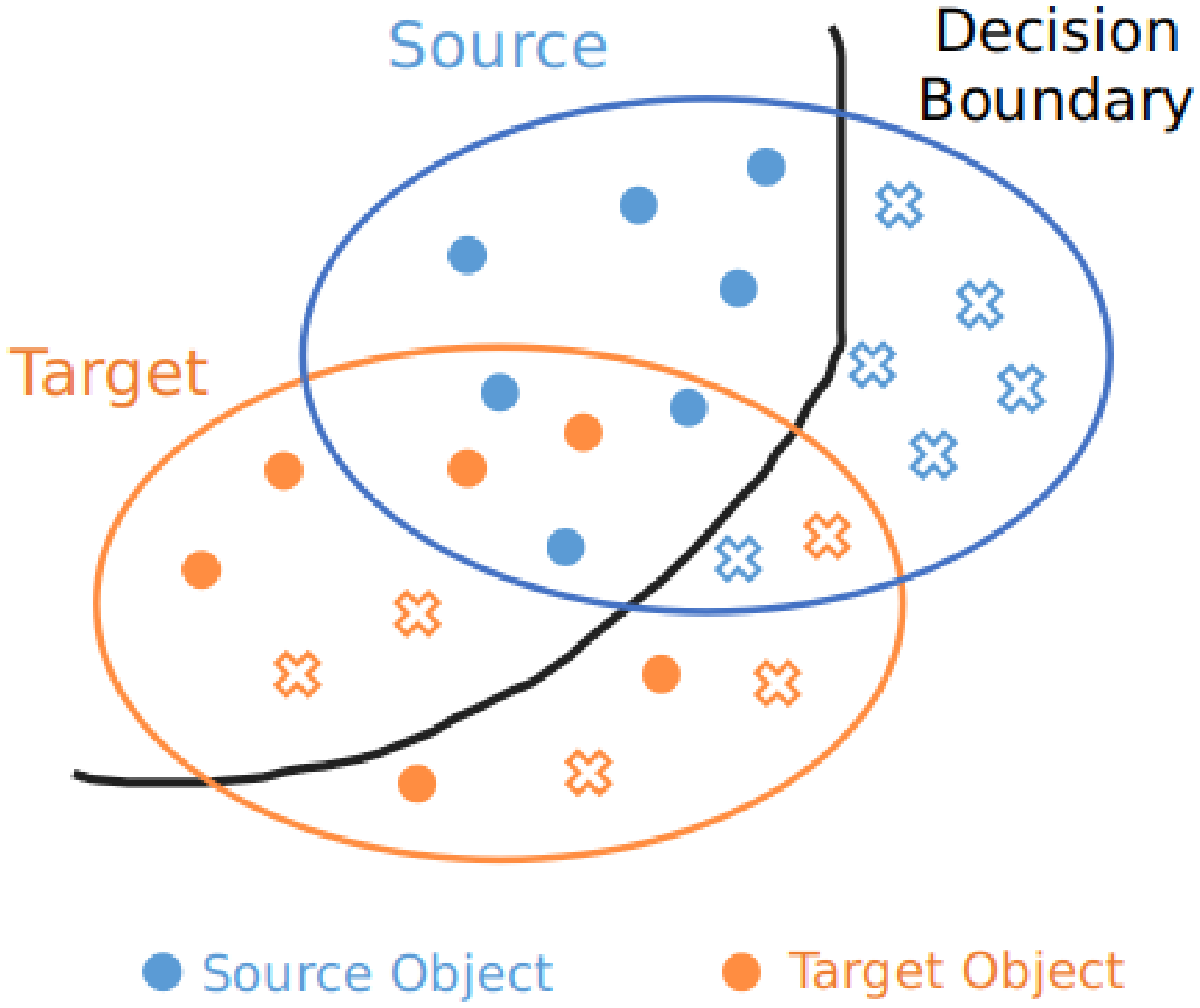}}
\subfigure[DA Faster~\cite{chen2018domain}]{
\label{fb_da_faster}
\includegraphics[width=.31\linewidth]{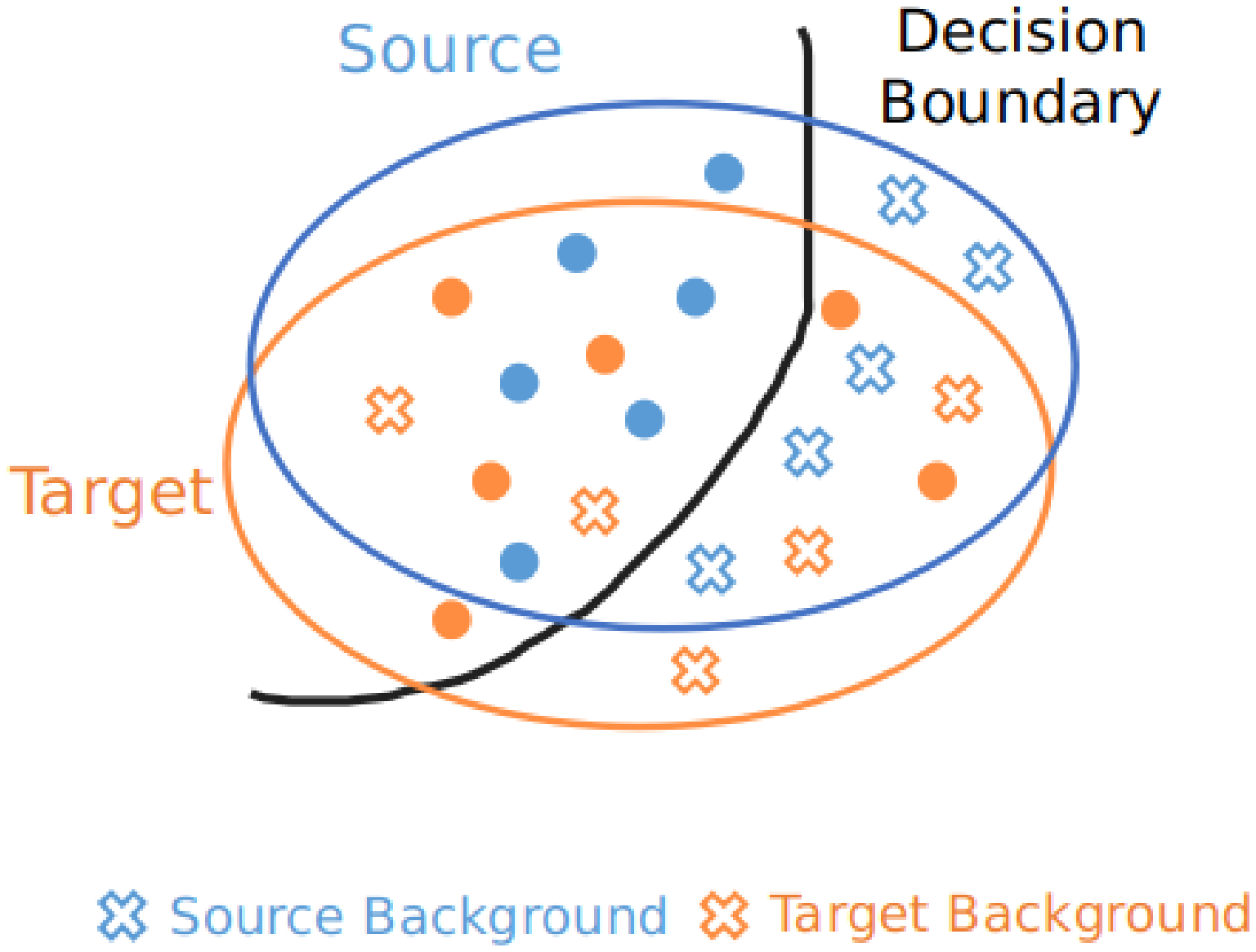}}
\subfigure[Ours]{
\label{fb_ours}
\includegraphics[width=.31\linewidth]{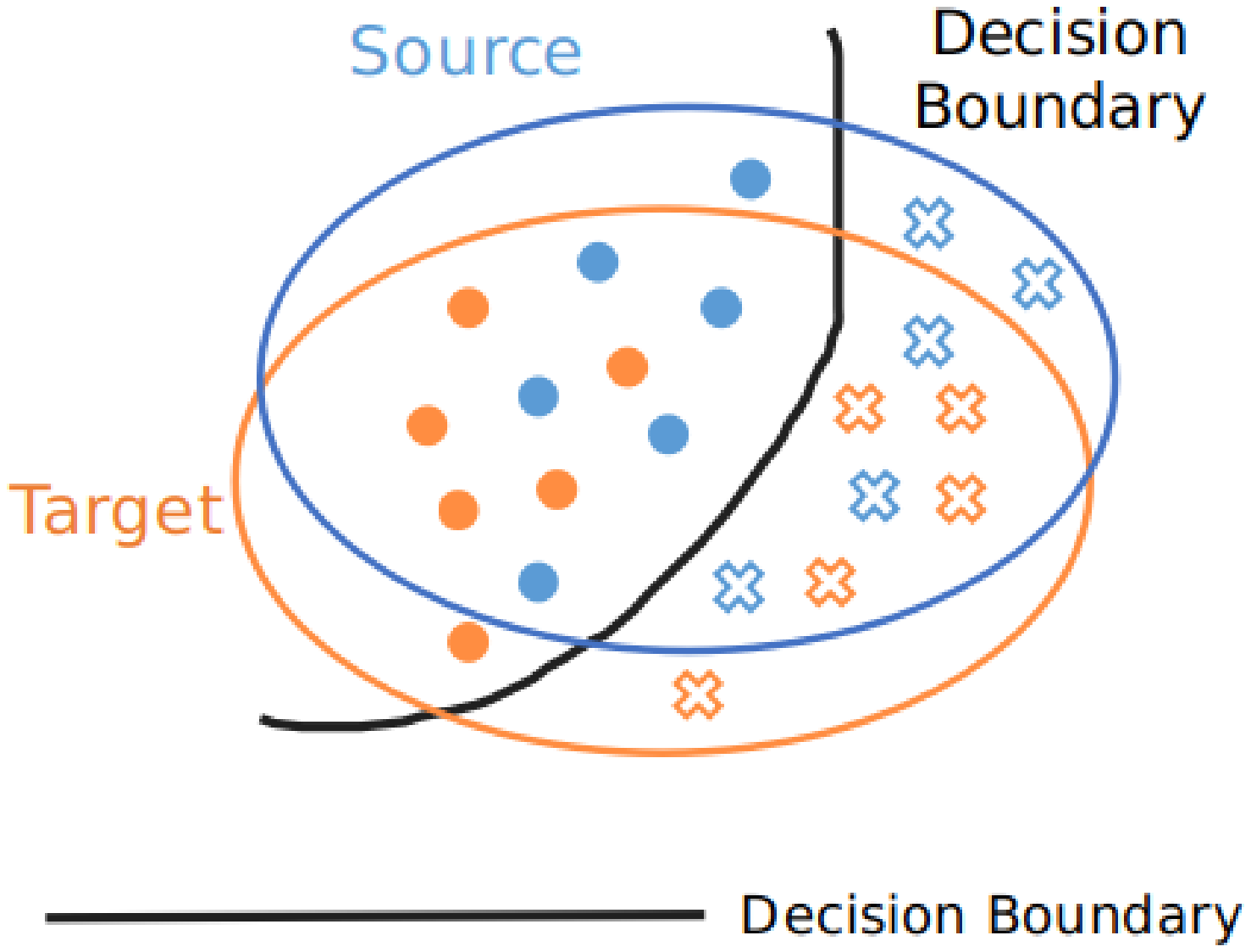}}
\caption{\small Illustration on different strategies for proposal-level feature alignment(best view in color). Although the domain gap is reduced successfully by DA Faster~\cite{chen2018domain}, there might exist misaligned foreground and background proposals between two domains. Our foreground-background aware alignment is expected to avoid such misalignment.}
\label{fig_fb}
\end{figure}

In cross-domain object detection, the background scenes can be massive and diverse, and the appearance of foreground objects also varies a lot. Simply matching the feature distributions of the instance-level features from ROI-pooling like~\cite{chen2018domain} may lead to misalignment between foreground and background regions between two domains, as the ROIs generated by RPN often contain many background regions. To this end, we propose to utilize two individual domain classifiers to separately align instance-level feature distributions for foreground and background proposals, respectively (see Fig.~\ref{fig_fb} for illustration).  

\def\cU{{U}}
\def\u{\mathbf{u}}
We also employ the adversarial training for instance-level distribution alignment. Let us denote by $D_{fg}$ and  $D_{bg}$ the object domain classifier and the background domain classifier, respectively. Given an image $x_i$, let us denote by $\cU = R(G(x_i))$ as the instance-level features after ROI-pooling using the region proposals generated from the RPN. We also use $\cU_{fg}$ and $\cU_{bg}$ to denote the sets of foreground and background instance-level features, respectively. 
The learning objective of proposal-level feature alignment can be written as:

\begin{equation}
\begin{split}
L_{roi}(x_i) = &\frac{1}{|\cU_{fg}|} \sum _{\u \in \cU_{fg}} L_{ce}\big(D_{fg}(\u), d_i\big) + \\ &\frac{1}{|\cU_{bg}|} \sum _{\u \in \cU_{bg}} L_{ce}\big(D_{bg}(\u), d_i\big)
\end{split}
\end{equation}

For the source domain, the foreground and background instances can be readily decided based on the ground-truth annotations. For the target domain, we select the instances with a prediction score larger than $0.9$ from the background (\textit{resp.}, any object) classifier as the background (\textit{resp.}, foreground) instances. 

\subsection{Overall Learning Objective}
The overall objective of our proposed approach is composed of mainly two parts, which are domain adaptation objective and object detection objective. The objective of domain adaptation can be written as:

\begin{equation}
\begin{split}
L_{da} = &\frac{1}{n_s} \sum_{i=1}^{n_s}\big( L_{feat}(x_i^s) + L_{roi}(x_i^s)\big)+ \\ &\frac{1}{n_t} \sum_{j=1}^{n_t}\big( L_{feat}(x_j^t) + L_{roi}(x_j^t)\big)
\end{split}
\end{equation}

Combining with detection loss, the overall objective is:
\begin{align}
\min_{G, R} \max_{D} L_{det}(G, R) - \lambda L_{da}(G, R, D)
\end{align}
where $\lambda$ is a leverage factor between $L_{da}$ and $L_{det}$, which can be optimized using standard back-propagation with the aid of GRL.

\section{Experiments}
In  this  section,  we  validate  our proposed DAA  model for cross-domain object detection with benchmark datasets.

\subsection{Experimental Setup}\label{exp:scenarios}
Following previous works~\cite{chen2018domain,inoue2018cross,saito2019strong}, we conduct experiments on four domain adaptation scenarios. We strictly keep the experimental settings the same as previous works unless otherwise specified.

\noindent\textbf{Cityscape to Foggy Cityscape~\cite{chen2018domain}:}
In this scenario, we aim to perform detection in foggy weather by adapting from the clear weather domain. The Cityscape dataset~\cite{Cordts_2016_CVPR} is used as the source domain and the Foggy-Cityscape dataset~\cite{sakaridis2018semantic} is used as the target domain. The VGG16~\cite{simonyan2014very} is adopted as the backbone network. 

\noindent\textbf{Pascal VOC to Clipart~\cite{inoue2018cross}:} This scenario is to adapt from the real images to artworks. The Pascal VOC dataset~\cite{everingham2010pascal} is taken as the source domain, and the  Clipart dataset~\cite{inoue2018cross}  is used as  the target domain, where the two datasets share 20 common categories. The Resnet101~\cite{He_2016_CVPR} pretrained in ImageNet~\cite{deng2009imagenet} is deployed as the backbone network.

\noindent\textbf{Pascal VOC to Watercolor~\cite{inoue2018cross}:} It also aims to adapt from the real images to artworks. The Pascal VOC dataset is used as the source, while the Watercolor dataset~\cite{inoue2018cross} with aesthetic style images is taken as the target domain, where two datasets share 6 common categories. The  Resnet101~\cite{He_2016_CVPR} pretrained in ImageNet~\cite{deng2009imagenet} is deployed as the backbone network.

\noindent\textbf{Sim10k to Cityscape~\cite{chen2018domain}:} In this scenario, the target is to leverage the abundant synthetic data for improving the performance of car detection in real world environment. The Sim10k dataset~\cite{johnson2016driving} is used as the source domain, in which 10k images drawn from video game Grand Theft Auto(GTA) with annotated labels are available to train the model. As the real domain, Cityscape~\cite{Cordts_2016_CVPR} is utilized to conduct domain adaptation task. These two datasets share a single common category of \textit{car}. In this experiment, the backbone network is VGG16~\cite{simonyan2014very}.

\subsection{Implementation details}
The detection part of our DAA is implemented by Faster R-CNN~\cite{ren2015faster}. In our experiments, we resize the short side of the image to 600 pixels during both training and testing phases. We use SGD with weight decay 0.0001 as the optimizing algorithm. The model is trained with learning rate of 0.001 in early 50,000 iterations and then is decayed to 0.0001 for another 20,000 iterations. We take $\lambda$ as 0.05 and $\eta$ as 5.0 to conduct the experiments except using $\lambda$=0.01 in Pascal VOC to Watercolor.  The batch size is set to 2 in common with previous works~\cite{he2019multi,saito2019strong,chen2018domain}. After all, the detection performance is reported after a total of 70,000 iterations with mean average precision(mAP) at IoU threshold of 0.5.

\subsection{Results}
We report the mAPs of our proposed DAA approach for four scenarios in Table~\ref{table:fog_compared}, \ref{table:clip_compared}, \ref{table:water_comapred} and \ref{table:car_comapred}, respectively. To validate the effectiveness of our DAA, the existing state-of-the-art methods reported on those four cases are also showed in the corresponding tables for comparison, including DA Faster R-CNN~\cite{chen2018domain}, SWDA~\cite{saito2019strong}, MAF~\cite{he2019multi}, PF~\cite{SHAN201931}, SCDA~\cite{Zhu_2019_CVPR}, DM~\cite{Kim_2019_CVPR}, WD~\cite{xu2019wasserstein} and MTOR~\cite{Cai_2019_CVPR}. The baseline "Source Only" which does not consider domain adaptation is also reported. We also include the "Oracle" results reported in~\cite{saito2019strong} for Cityscape to Foggy Cityscape and SIM10K to Cityscape, which uses the target training split with ground truth annotations to train the model.

We observe that our DAA model outperforms existing state-of-the-art (SOTA) methods for all scenarios, except on SIM10K to Cityscape where our DAA model is slightly worse than the SCDA method by $1.6\%$. In particular, as shown in Table.~\ref{table:fog_compared}, on the scenario of Cityscape to Foggy Cityscape, our DAA outperforms previous SOTA MTOR~\cite{Cai_2019_CVPR} by a good margin of $4.1\%$. It is also worthy mentioning that the performance of our DAA approach is very close to the "Oracle" result, showing the superior domain alignment ability of our DAA model on this scenario. For the real to artistic adaptation, our DAA model also surpasses the previous SOTAs, and produces new SOTA performance of $42.3\%$ and $55.1\%$ on Clipart and Watercolor, respectively. These observations clearly demonstrate the effectiveness of our DAA approach by deeply aligning the domain distributions.

\setlength{\tabcolsep}{4pt}
\begin{table}[t]
\small
\centering
\caption{The Average Precisions (\%) of different methods on the scenario of Cityscape$\rightarrow$Foggy Cityscape. The best results are denoted in bold.}
\label{table:fog_compared}
\renewcommand\tabcolsep{0.9pt}
\begin{tabular}{c|cccccccc|c}
\hline
Method & bus & bicycle & car & mcycle &person & rider & train & truck& mAP\\
\hline
\hline
Source Only  & 23.8 &24.7&  27.2 &14.2 &23.1 & 30.3 &9.1 & 12.2& 20.6 \\
\hline
DA-Faster  & 35.3& 27.1   &  40.5& 20.0& 25.0& 31.0& 20.2& 22.1& 27.6\\
\hline
PF~\cite{SHAN201931} & -&- &-  &- & -&- &- &- & 28.9\\
\hline
SCDA~\cite{Zhu_2019_CVPR} & 39.0& 33.6& 48.5& 28.0&  \textbf{33.5}& 38.0& 23.3& 26.5& 33.8\\
\hline
DM~\cite{Kim_2019_CVPR}& 38.4& 32.2& 44.3& 28.4& 30.8& 40.5& 34.5& 27.2& 34.6\\
\hline
MAF~\cite{he2019multi} & 39.9& 33.9& 43.9& 29.2& 28.2& 39.5& 33.3& 23.8& 34.0\\
\hline
WD~\cite{xu2019wasserstein} & 39.9& 34.4& 44.2& 25.4& 30.2& 42.0& 26.5& 22.2& 33.1\\
\hline
MTOR~\cite{Cai_2019_CVPR} & 38.6& 35.6& 44.0& 28.3& 30.6& 41.4& \textbf{40.6}& 21.9& 35.1\\
\hline
SWDA~\cite{saito2019strong} & 36.2& 35.3& 43.5& 30.0& 29.9& 42.3& 32.6& 24.5& 34.3\\
\hline
\hline
Ours &    \textbf{46.6}&  \textbf{36.9} &\textbf{48.8}  & \textbf{34.0}& 33.2& \textbf{47.6}& 38.2& \textbf{28.1} & \textbf{39.2}\\
\hline
\hline
Oracle  &     50.0& 36.2& 49.7& 34.7& 33.2& 45.9& 37.4& 35.6& 40.3\\
\hline
\end{tabular}
\end{table}

\begin{table*}[t]
\centering
\renewcommand\tabcolsep{0.8pt}
\caption{The Average Precisions (\%) of different methods on the scenario of Pascal VOC$\rightarrow$Clipart. The best results are denoted in bold.}
\label{table:clip_compared}
\resizebox{\textwidth}{!}{
\begin{tabular}{c|cccccccccccccccccccc|c}
\hline
Method  &  aero &bcycle & bird & boat & bottle & bus & car & cat & chair & cow & table & dog & hrs & mbike & prsn & plnt & sheep & sofa & train & tv & mAP\\
\hline
\hline
Source Only  &   29.5& 49.1& 28.3& 17.6& 27.9& 34.2& 26.3& 12.4& 26.9& 30.9& 13.3& 9.3& 21.0& 54.2& 45.3& 33.7& 11.6& 20.9& 20.0& 36.9& 27.5 \\
\hline
DA-Faster~\cite{chen2018domain} & 24.1 &  41.9 & 28.2  & 24.6 & 33.9  &    39.9& 29.7 &10.5 &33.7 &24.1 & 12.7& 10.8& 25.5&51.0 & 43.1& 37.1& 11.8& 24.4& 39.2& 42.1& 29.4\\
\hline
DM~\cite{Kim_2019_CVPR} & 25.8&  \textbf{63.2}& 24.5& \textbf{42.4}& \textbf{47.9}& 43.1& 37.5& 9.1& \textbf{47.0}& 46.7& \textbf{26.8}& \textbf{24.9}& \textbf{48.1}& 78.7& 63.0& 45.0& \textbf{21.3}& \textbf{36.1}& 52.3& \textbf{53.4}& 41.8\\
\hline
MAF~\cite{he2019multi} & 32.1& 52.2 & 33.2& 31.1 & 42.6 & 38.7 & 36.8 & 15.7 &37.0 & 51.3& 20.8 & 16.0 &35.2 & 64.7 & 60.4 & 47.3 & 18.7 & 27.6 & 47.9 & 42.8 & 37.6\\
\hline
SWDA~\cite{saito2019strong}& 26.2&  48.5& 32.6& 33.7&38.5 & 54.3& 37.1& 18.6& 34.8& 58.3& 17.0& 12.5& 33.8& 65.5& 61.6& \textbf{52.0}& 9.3& 24.9& 54.1& 49.1& 38.1\\
\hline
\hline
Ours   &     \textbf{35.0}& 59.5& \textbf{34.6}& 30.2& 38.1&\textbf{60.2}& \textbf{40.2}& \textbf{20.5}& 39.3& \textbf{58.5}& 26.4& 22.8& 33.8& \textbf{82.9}& \textbf{64.4}& 48.8& 18.0& 28.6& \textbf{57.6}& 46.2& \textbf{42.3}\\
\hline
\end{tabular}
} 
\vspace{-10pt}
\end{table*}

\begin{table}[t]
\centering
\renewcommand\tabcolsep{0.9pt}
\caption{The Average Precisions (\%) of different methods on the scenario of Pascal VOC$\rightarrow$Watercolor. The best results are denoted in bold.}
\label{table:water_comapred}
\begin{tabular}{c|cccccc|c}
\hline
Method  & bcycle & bird & car & cat &  dog & person & mAP\\
\hline
\hline
Source Only  &  66.1& 40.7& 44.1& 33.2& 30.4& 63.9& 46.4\\
\hline
DA-Faster~\cite{chen2018domain} & 62.9 & 51.4 & 48.5 & 35.3 & 25.7 & 60.2 & 47.3 \\
\hline
DM~\cite{Kim_2019_CVPR}& -&- &- &- &- &-  & 52.0\\
\hline
MAF~\cite{he2019multi}& 72.9  & 55.8 & \textbf{50.3} & 38.7 &  35.7&  65.0  & 53.1\\
\hline
SWDA~\cite{saito2019strong} &   82.3& 55.9 & 46.5 & 32.7 & 35.5 & \textbf{66.7} & 53.3\\
\hline
\hline
Ours  &  \textbf{85.1}&\textbf{56.6} &46.2  & \textbf{39.9}&  \textbf{36.9}& 65.6 & \textbf{55.1}\\
\hline
\end{tabular}
\vspace{-10pt}
\end{table}


\begin{table}[t]
  \centering
 \caption{Average Precisions(\%) of different methods on the scenario of Sim10K$\rightarrow$Cityscape. The best result is denoted in bold.}
\label{table:car_comapred}
\begin{tabular}{c|c}
\hline
Method & AP on car\\
\hline
\hline
Source Only& 35.5\\
DA-Faster~\cite{chen2018domain}&38.9\\
PF~\cite{SHAN201931}& 39.6\\
SCDA~\cite{Zhu_2019_CVPR}& \textbf{43.0}\\
MAF~\cite{he2019multi}& 41.1\\
WD~\cite{xu2019wasserstein}& 40.6\\
SWDA~\cite{saito2019strong}& 40.1\\
\hline
\hline
Ours& 41.4\\
\hline
\hline
Oracle& 53.1\\
\hline
\end{tabular}
\vspace{-10pt}
\end{table}

\subsection{Experimental Analysis}\label{subsec:analysis}
We further conduct an ablation study  by switching on/off different modules in our DAA model to validate their effects. We take the scenario of Cityscape~\cite{Cordts_2016_CVPR} to Foggy Cityscape~\cite{sakaridis2018semantic} as an example to analyze, and the results on the other datasets are included in the Supplementary. 

In particular, we use "L", "T", "G", "M", "I" to represent the local adaptor, the transition adaptor, the global adaptor, the domain mask integration module, and the instance-level foreground-background aware alignment, respectively. The mAPs of of different special cases of our DAA model by switching on/off certain modules are reported in Table~\ref{table:fog_ablation}. We have the following observations. 

\noindent\textbf{Transition Adaptor:} We first validate the effectiveness of our transition adaptor. As shown in  Table~\ref{table:fog_ablation}, our special case DAA-A which uses only the local adaptor (L) and global adaptor (G) gives an mAP of $27.1\%$, while another special case DAA-B which additionally uses the transition adaptor (T) gives $37.0\%$. Similarly, if we further taking into account the domain mask integration module (M), DAA-D improves DAA-C from $33.9\%$ to $37.6\%$ by additionally using the transition adaptor (T). These results confirm that it is important to bridge the local and global alignment with a transition adaptor to cope with the gradually shifting transferability of features in CNNs.

\begin{figure}[t]
\centering
 \subfigure[S w/o DMI]{
\centering
 \includegraphics[width=.22\linewidth]{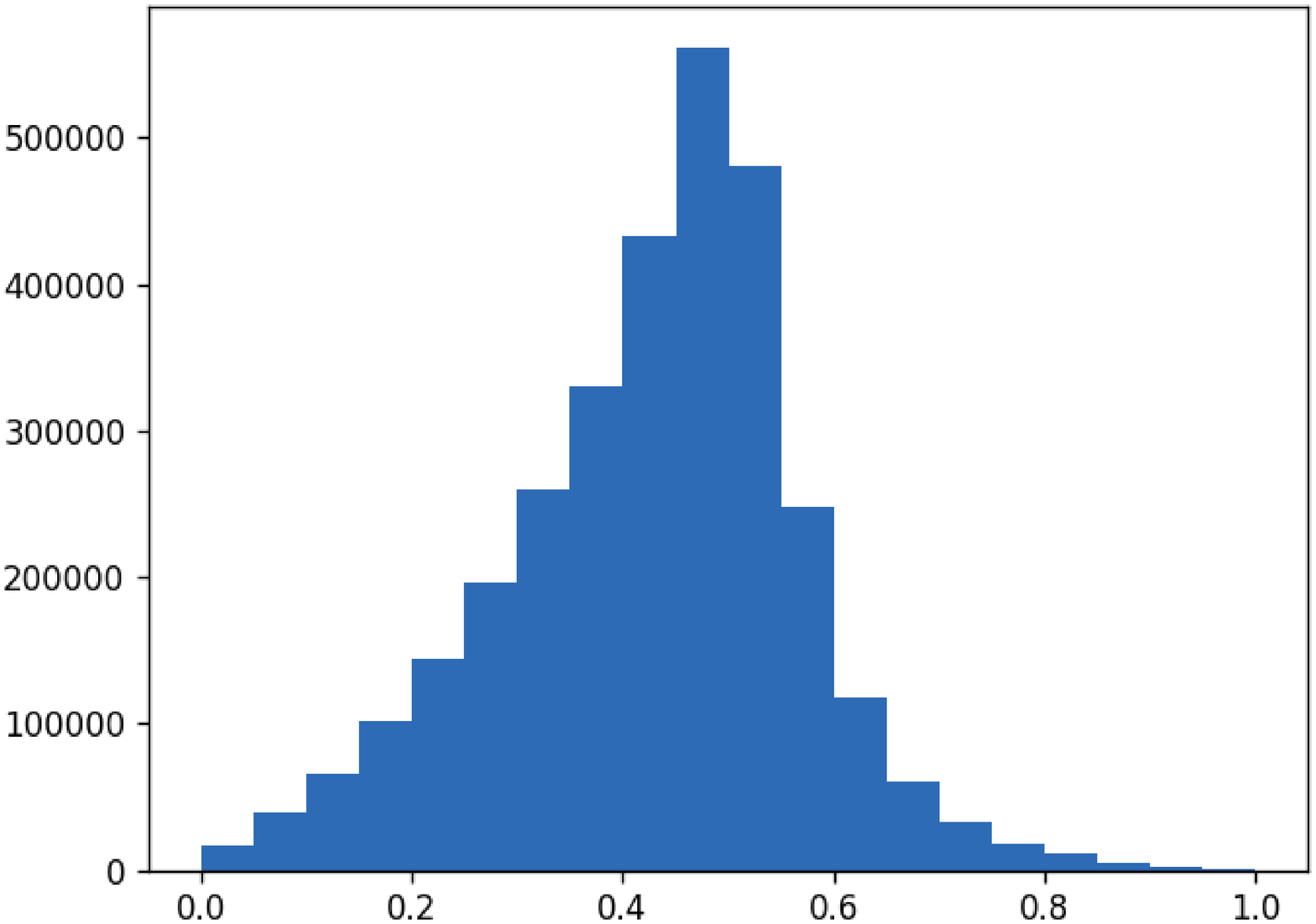}
\label{hist_source_wo_mask}}
\subfigure[S w DMI]{
    \centering
   \includegraphics[width=.22\linewidth]{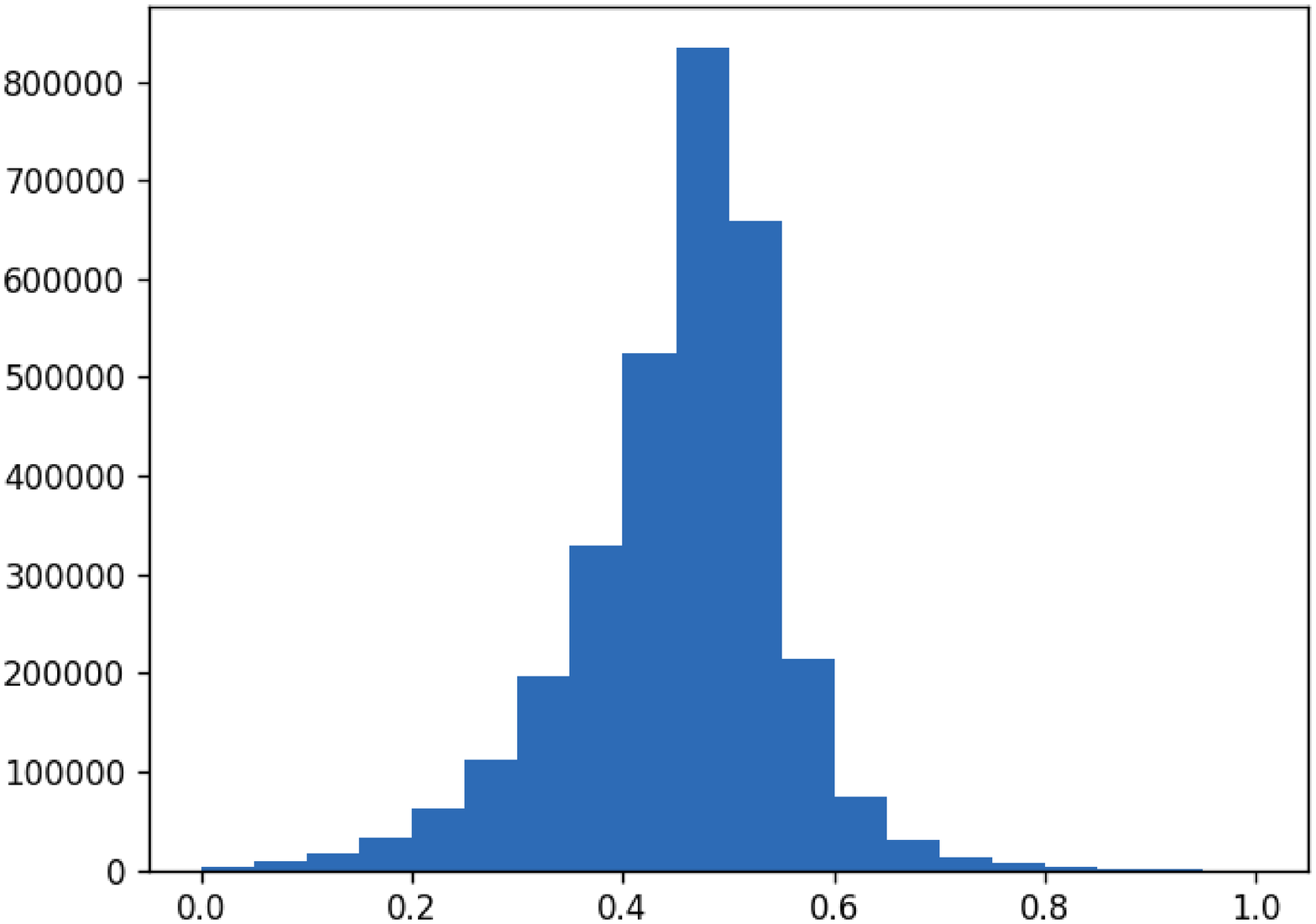}
    \label{hist_source_with_mask}}
  \subfigure[T w/o DMI]{
    \centering
   \includegraphics[width=.22\linewidth]{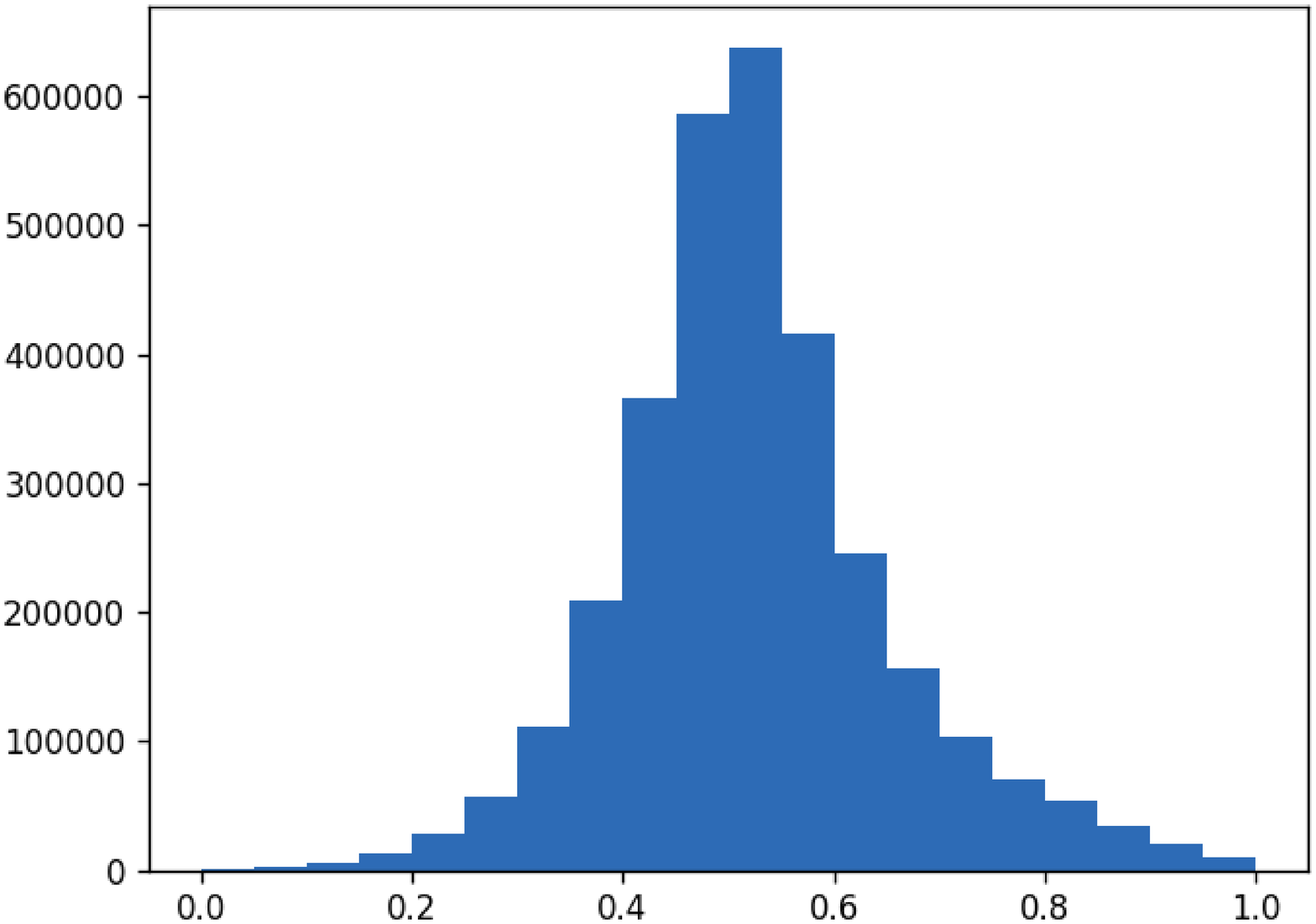}
    \label{hist_target_wo_mask}}
  \subfigure[T w DMI]{
    \centering
    \includegraphics[width=.22\linewidth]{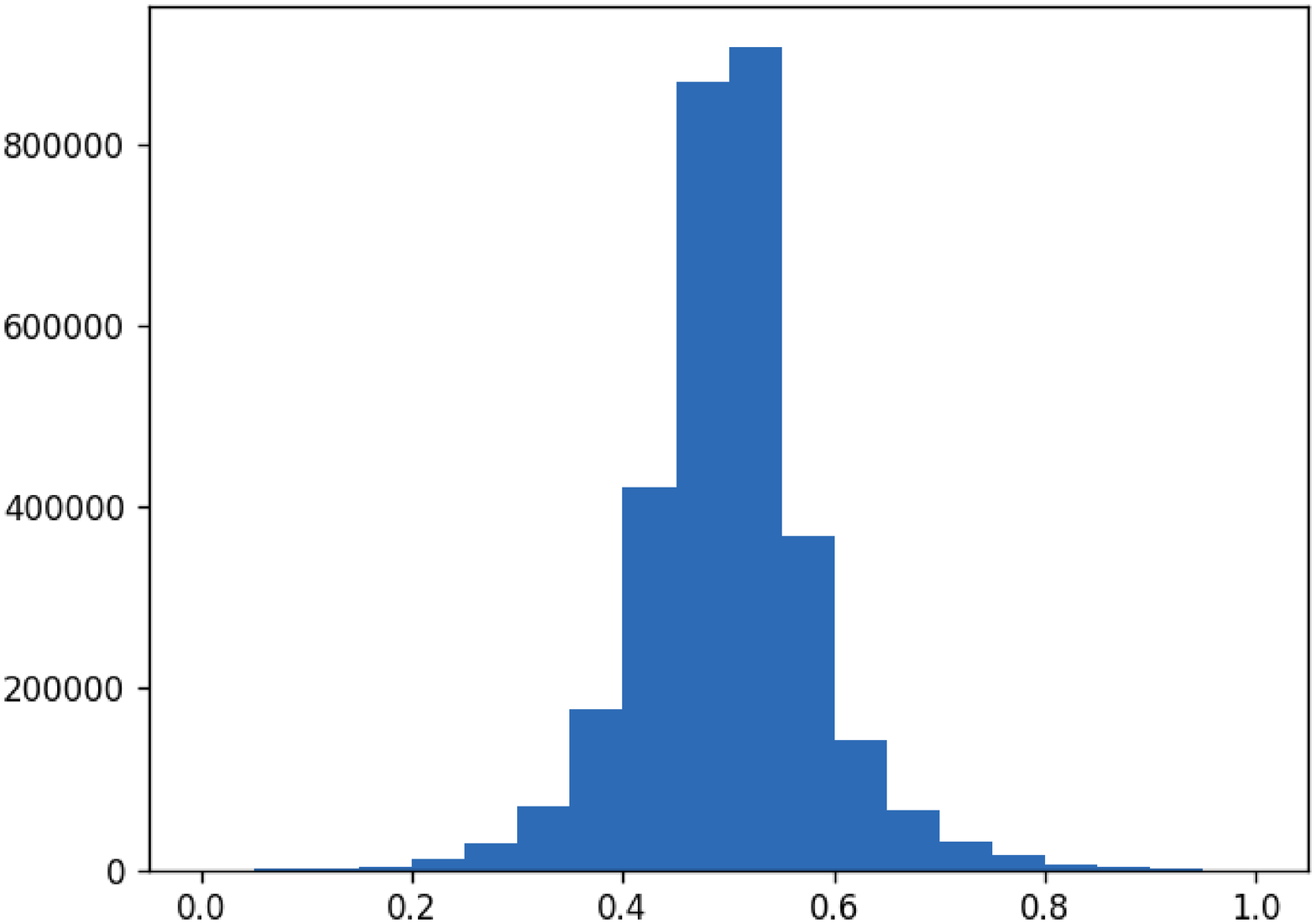}
    \label{hist_target_w_mask}}
  \caption{\small Frequency distribution histogram of local features generated by $G_1$, in which $S$ and $T$ denote features are drawn from the source or target domain, respectively. The mark of $w$ stands taking DMI during training while $w/o$ does not.}
  \label{hist_pic}
\end{figure}

\noindent\textbf{Domain Mask Integration:} We then investigate the domain mask integration (DMI) module. By comparing DAA-C with DAA-A, and DAA-D with DAA-B we can clearly observe that the DMI module is helpful for improving the domain adaptation. Also, we further include a special case of SWDA for comparison (denoted as SWDA* in Table~\ref{table:fog_ablation}), in which we keep only its strong local and weak global alignment modules. We observe that our DAA-C achieves better mAP than SWDA*, which confirms our discussion in Section~\ref{DMI} that strengthening the local alignment as in our DAA would be more effective than relaxing the global alignment as in SWDA to model different transferability for features of bottom and top layers. Moreover, we also depict the statistics on the domain probability scores of source/target samples  with or without using DMI in Fig~\ref{hist_pic}. We observe that when using DMI, the domain probabilities are more centered around $0.5$, which indicates the adversarial training reaches a better convergence point. Namely, the features cannot be easily distinguished, and thus  more likely to be more domain invariant.

\noindent\textbf{Foreground-background Aware Alignment:} Finally, we evaluate our proposed foreground-background aware alignment. By appending this module, our final DAA model improves the DAA-D from $37.6\%$ to $39.2\%$, which clearly validates the importance of handling the object variance with instance-level alignment for cross-domain detection. Moreover, if we do not distinguish the foreground and background in alignment as in \cite{chen2018domain}, the performance drops to $36.9\%$, which proves the effectiveness for our new  foreground-background aware alignment module. Qualitative results are provided in the Supplementary.  

\setlength{\tabcolsep}{4pt}
\begin{table}[t]
\begin{center}
\caption{\small Ablation study on our proposed DAA approach. 
}
\label{table:fog_ablation}
\renewcommand\tabcolsep{0.8pt}
\resizebox{1\linewidth}{!}{
\begin{tabular}{c|ccccc|ccccccccc}
\hline
Method & L & T & G & M & I & bus & bicycle & car & mcycle &person & rider & train & truck& mAP\\
\hline
SWDA* &  \checkmark &  &\checkmark&  &&      35.3& 32.0& 43.8& 24.2& 29.4& 40.6& 26.0& 22.5& 31.7\\
\hline
DAA-A  &  \checkmark &  &\checkmark  && & 33.3 & 26.1 &35.7 & 22.4& 25.4 & 34.0 & 21.6& 18.3& 27.1\\
DAA-B  &  \checkmark &  \checkmark&\checkmark  & &&\textbf{48.1} &35.1 & 48.4 & 29.6& 33.0 & 45.7& 27.3& 29.1& 37.0\\
DAA-C  &  \checkmark &  &\checkmark  &\checkmark &&   37.3& 35.0& 44.0& 26.4& 31.3& 44.3& 27.4& 25.2& 33.9\\
DAA-D  &  \checkmark &\checkmark  &\checkmark  &\checkmark & &45.2  & \textbf{37.4}&   \textbf{49.1}& 32.0&33.0 &47.3 & 27.1 & \textbf{30.0} & 37.6\\
\hline
DAA  &  \checkmark &  \checkmark&\checkmark  & \checkmark&\checkmark&  46.6&  36.9 &48.8  & \textbf{34.0}&  \textbf{33.2}& \textbf{47.6}& \textbf{38.2}& 28.1 & \textbf{39.2}\\
\hline
\end{tabular}}
\end{center}
\vspace{-10pt}
\end{table}

\section{Conclusion}
In this work, we propose a novel Faster R-CNN based method for object detection under the unsupervised domain adaptation setting. Considering neural network layers have different transferabilities, we introduce different alignment strategies for those layers in order to effectively reduce the mismatch between the source and target domains. Moreover, after obtaining region proposals in our method, a newly developed foreground-background aware alignment module is further performed to alleviate the misalignment between foreground and background regions from the two domains. We conduct extensive experiments on several benchmark datasets, and the results clearly demonstrate that our method achieves superior performance over the existing state-of-the-art baselines.

{\small
\bibliographystyle{ieee_fullname}
\bibliography{egbib}
}

\onecolumn
\setcounter{section}{0}
\section*{Supplemental Material}
\setcounter{section}{0}

In this Supplementary, we additionally provide: 1) ablation study on all datasets; 2) qualitative comparison with state-of-the-arts; and 3) network architecture of different discriminators. 

\section{Additional Results}
In Section~\ref{subsec:analysis} of the main paper, we have used the scenario of Cityscape to Foggy Cityscape as an example to conduct ablation study on our DAA approach, and validated the effectiveness of different components by gradually switching on/off them. In this section,  we additionally provide the ablation study results on the  other three scenarios. Moreover, in order to validate the effectiveness of our foreground-background aware alignment, we further conduct an ablation study by replacing this component with the ordinary feature alignment as presented in DA-Faster, \emph{i.e.}, deploying only one domain discriminator on all proposals. We denote this special case as \textit{DAA-E}. The results are shown in Table~\ref{table:ab_clip},~\ref{table:ab_water},~\ref{table:ab_car},~\ref{table:ab_fog}, where we generally have similar observations as in the main paper. 

\setlength{\tabcolsep}{4pt}
\begin{table}[!htbp]
\centering
\caption{\small Ablation study on Pascal VOC to Clipart. $\dagger$ denotes replacing object and background discriminators with only one domain discriminator in instance level.}
\label{table:ab_clip}
\renewcommand\tabcolsep{0.9pt}
\resizebox{\textwidth}{!}{
\begin{tabular}{c|ccccc|ccccccccccccccccccccc}
\hline
Method & L & T & G & M & I & aero &bcycle & bird & boat & bottle & bus & car & cat & chair & cow & table & dog & hrs & mbike & prsn & plnt & sheep & sofa & train & tv & mAP\\
\hline
SWDA* & \checkmark &  &\checkmark& & &      30.7& 53.9& 30.3& 23.2& 37.9& 58.8& 30.1& 18.0& 37.6& 33.8& 22.9& 18.2& \textbf{36.1}& 74.1& 58.1& 45.1& 14.2& 22.2& 42.7& 48.3& 36.8\\
\hline
DAA-A &  \checkmark &  &\checkmark  & &&   30.0& 39.5& 23.8& 29.7 &30.3 &35.2 & 36.2& 10.9& 33.7& 41.5& 18.8& 12.7& 30.2& 40.8& 48.6& 41.1 & 10.3& 25.1 & 34.9& 37.4 & 30.5\\
DAA-B  &  \checkmark &  \checkmark&\checkmark  & &&     \textbf{36.7}& 52.5& 36.3& 27.4& 38.5& 44.1& 36.7& 16.5& 36.3& 58.5& 16.1& \textbf{27.2}& 35.5& 71.7& 59.3& 40.2& 11.0 & 21.8& 44.3& 40.1& 37.5\\
DAA-C  &  \checkmark &  &\checkmark  &\checkmark &&       34.0& 51.3& 25.2& \textbf{31.0}& \textbf{39.7}& 35.0& 35.6& 6.2& 28.3& 48.7& 23.1& 16.0& 30.8& 59.4& 57.5& \textbf{50.5}& 20.5& 23.4& 47.1& 40.2& 35.2\\
DAA-D  &  \checkmark &\checkmark  &\checkmark  &\checkmark & &34.5& 58.0& \textbf{36.6}&28.2 & 30.8& 56.2 & \textbf{41.0}& 13.1& 39.7& \textbf{59.4}& 19.8& 19.5& 31.5& 70.9& 63.6& 41.8& 15.7& \textbf{28.8}&51.7 &\textbf{50.3}& 39.6\\
DAA-E & \checkmark&\checkmark& \checkmark& \checkmark& \checkmark$\dagger$& 33.5& 49.3& 28.5& 30.7& 38.0& 57.4& 39.5& 11.5& \textbf{42.0}& 56.6& 23.5& 20.0 & 30.0& 56.6& 58.8& 45.5& \textbf{21.5}& 26.8& 40.8& 48.4& 38.0\\
\hline
DAA &  \checkmark &  \checkmark&\checkmark  & \checkmark&\checkmark &     35.0& \textbf{59.5}& 34.6& 30.2& 38.1&\textbf{60.2}& 40.2& \textbf{20.5}& 39.3& 58.5& \textbf{26.4}& 22.8& 33.8& \textbf{82.9}& \textbf{64.4}& 48.8& 18.0& 28.6& \textbf{57.6}& 46.2& \textbf{42.3}\\
\hline
\end{tabular}}
\end{table}
\setlength{\tabcolsep}{1.4pt}

\setlength{\tabcolsep}{4pt}
\begin{table}[!htbp]
\centering
\begin{minipage}[t]{.5\linewidth}
\centering
\caption{\small Ablation study on Pascal VOC to Watercolor.}
\label{table:ab_water}
\renewcommand\tabcolsep{0.9pt}
\begin{tabular}{c|ccccc|ccccccc}
\hline
Method & L & T & G & M & I  & bcycle & bird & car & cat &  dog & person & mAP\\
\hline
SWDA* &  \checkmark &  &\checkmark& & &          71.1& 52.0& 49.7& 34.5& 36.8& 64.9& 51.5\\
\hline
DAA-A&  \checkmark &  &\checkmark  & &&       83.3 & 49.3 & 48.6 & 33.6 & 28.8 & 60.8 & 50.7\\
DAA-B&  \checkmark &  \checkmark&\checkmark  && &          84.7&51.5 & \textbf{52.0} & 31.3 & 34.7 & 63.8 & 53.0\\
DAA-C&  \checkmark &  &\checkmark&\checkmark& &         81.1& 54.2& 48.5& 32.7& 33.2& 64.0& 52.3 \\
DAA-D&  \checkmark &\checkmark  &\checkmark  &\checkmark && 84.0  & 51.0&  49.5& 39.8 &33.6 & 64.2 & 53.7\\
DAA-E&  \checkmark &\checkmark  &\checkmark  &\checkmark &\checkmark$\dagger$& 77.7& 53.0& 45.5& 38.8& 35.5& 62.5& 52.2\\
\hline
DAA &  \checkmark &\checkmark  &\checkmark  &\checkmark &\checkmark& \textbf{85.1}&\textbf{56.6} &46.2  & \textbf{39.9}&  \textbf{36.9}& \textbf{65.6} & \textbf{55.1}\\
\hline
\end{tabular}
\end{minipage}
\begin{minipage}[t]{.45\linewidth}
\scriptsize
\begin{center}
\caption{\small Ablation study on Sim10k to Cityscape.}
\label{table:ab_car}
\renewcommand\tabcolsep{0.9pt}
\resizebox{.6\textwidth}{!}{
\begin{tabular}{c|ccccc|c}
\hline
Method & L & T & G & M & I  & AP on car\\
\hline
SWDA* &  \checkmark &  &\checkmark && &38.4\\
\hline
DAA-A&  \checkmark &  &\checkmark  && & 33.9\\
DAA-B&  \checkmark &  \checkmark&\checkmark  & &&41.0\\
DAA-C&  \checkmark &  &\checkmark  &\checkmark &&40.7\\
DAA-D&  \checkmark &\checkmark  &\checkmark  &\checkmark & &41.2\\
DAA-E&  \checkmark &\checkmark  &\checkmark  &\checkmark &\checkmark$\dagger$ & 38.2\\
\hline
DAA  &  \checkmark &\checkmark  &\checkmark  &\checkmark &\checkmark & \textbf{41.4}\\
\hline
Oracle  &  &  &  & & &     53.1\\
\hline
\end{tabular}}
\end{center}
\end{minipage}
\end{table}
\setlength{\tabcolsep}{1.4pt}

\setlength{\tabcolsep}{4pt}
\begin{table}[!htbp]
\begin{center}
\caption{\small Ablation study on Cityscape to Foggy Cityscape.} 
\label{table:ab_fog}
\renewcommand\tabcolsep{0.9pt}
\resizebox{.6\textwidth}{!}{
\begin{tabular}{c|ccccc|ccccccccc}
\hline
Method & L & T & G & M& I  & bus & bicycle & car & mcycle &person & rider & train & truck& mAP\\
\hline
SWDA* &  \checkmark &  &\checkmark&  &&        35.3& 32.0& 43.8& 24.2& 29.4& 40.6& 26.0& 22.5& 31.7\\
\hline
DAA-A  &  \checkmark &  &\checkmark  &&& 33.3 & 26.1 &35.7 & 22.4& 25.4 & 34.0 & 21.6& 18.3& 27.1\\
DAA-B  &  \checkmark &  \checkmark&\checkmark  & && \textbf{48.1} &35.1 & 48.4 & 29.6& 33.0 & 45.7& 27.3& 29.1& 37.0\\
DAA-C  &  \checkmark &  &\checkmark  &\checkmark &&   37.3& 35.0& 44.0& 26.4& 31.3& 44.3& 27.4& 25.2& 33.9\\
DAA-D  &  \checkmark &\checkmark  &\checkmark  &\checkmark & &45.2  & \textbf{37.4}&   \textbf{49.1}& 32.0&33.0 &47.3 & 27.1 & \textbf{30.0} & 37.6\\
DAA-E &  \checkmark &\checkmark  &\checkmark  &\checkmark &\checkmark$\dagger$ & 46.3& 36.4& 45.0& 32.4& 33.1& 45.4& 29.4& 27.1& 36.9\\
\hline
DAA &  \checkmark &\checkmark  &\checkmark  &\checkmark &\checkmark& 46.6&  36.9 &48.8  & \textbf{34.0}& \textbf{33.2}& \textbf{47.6}& \textbf{38.2}& 28.1 & \textbf{39.2}\\
\hline
Oracle  &  &  &  && &     50.0& 36.2& 49.7& 34.7& 33.2& 45.9& 37.4& 35.6& 40.3\\
\hline
\end{tabular}}
\end{center}
\end{table}
\setlength{\tabcolsep}{1.4pt}

\section{Qualitative Results}
\begin{figure}[!htbp]
\centering
  \subfigure[GT]{
  \begin{minipage}[t]{.2\textwidth}
    \centering
   \includegraphics[width=1\linewidth]{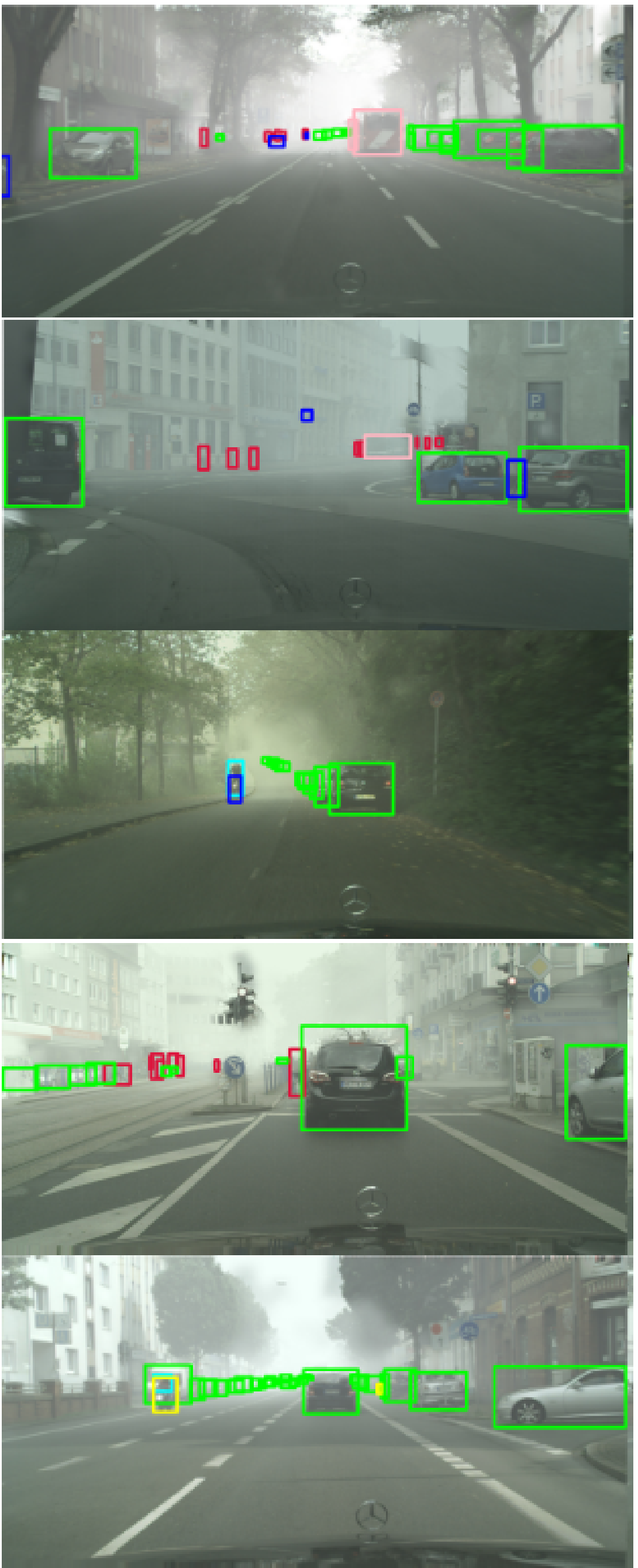}
    \label{cam_fog_anno}
    \end{minipage}}
\hspace{-8pt}
    \subfigure[Source Only]{
  \begin{minipage}[t]{.198\textwidth}
    \centering
   \includegraphics[width=1\linewidth]{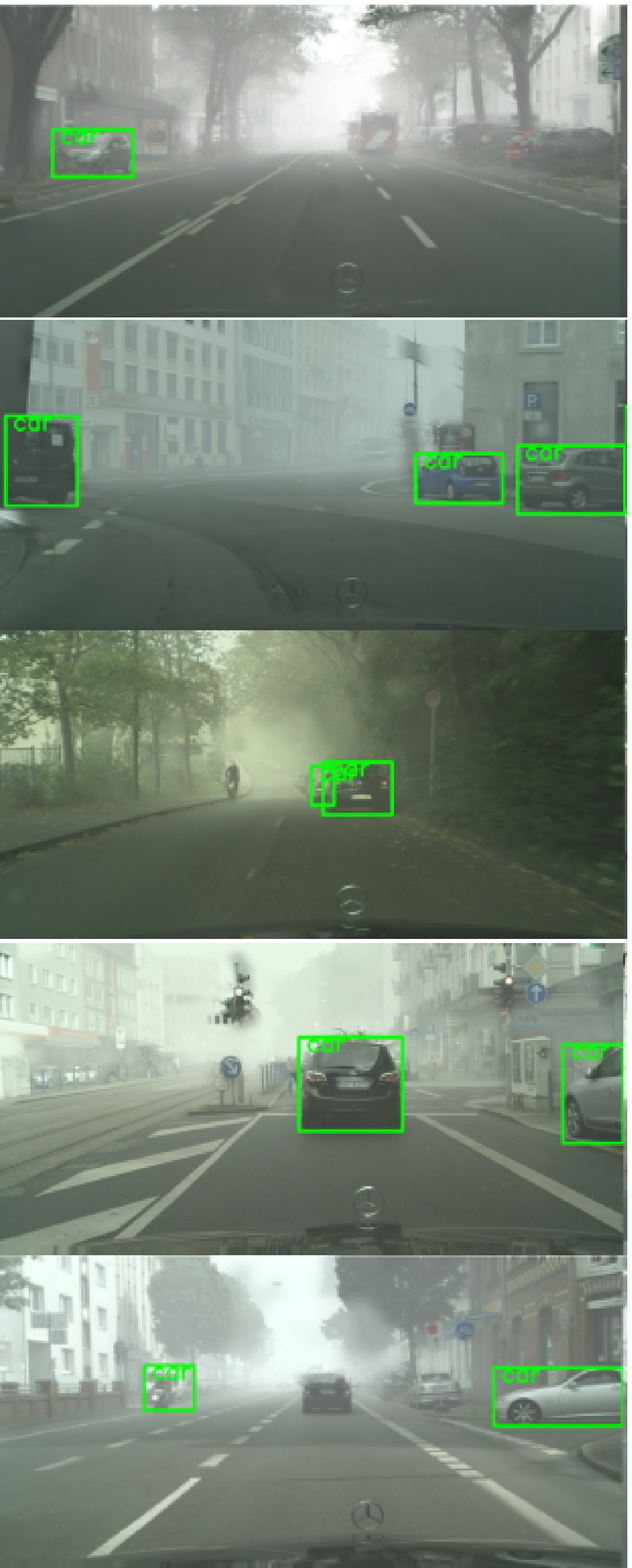}
    \label{cam_fog_source_only}
    \end{minipage}}
\hspace{-10pt}
  \subfigure[DA-Faster]{
  \begin{minipage}[t]{.2\textwidth}
    \centering
   \includegraphics[width=1\linewidth]{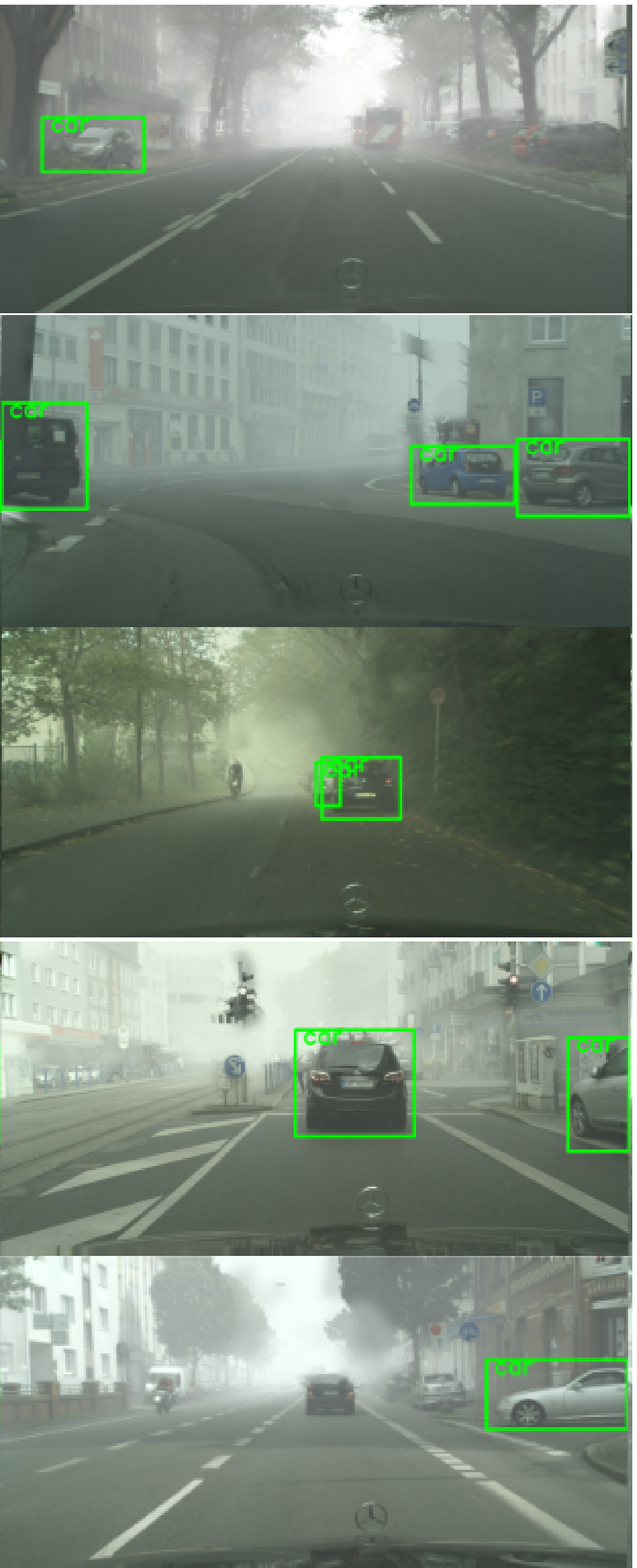}
    \label{cam_fog_da_faster}
    \end{minipage}}
\hspace{-9pt}
  \subfigure[SWDA]{
  \begin{minipage}[t]{.199\textwidth}
    \centering
    \includegraphics[width=1\linewidth]{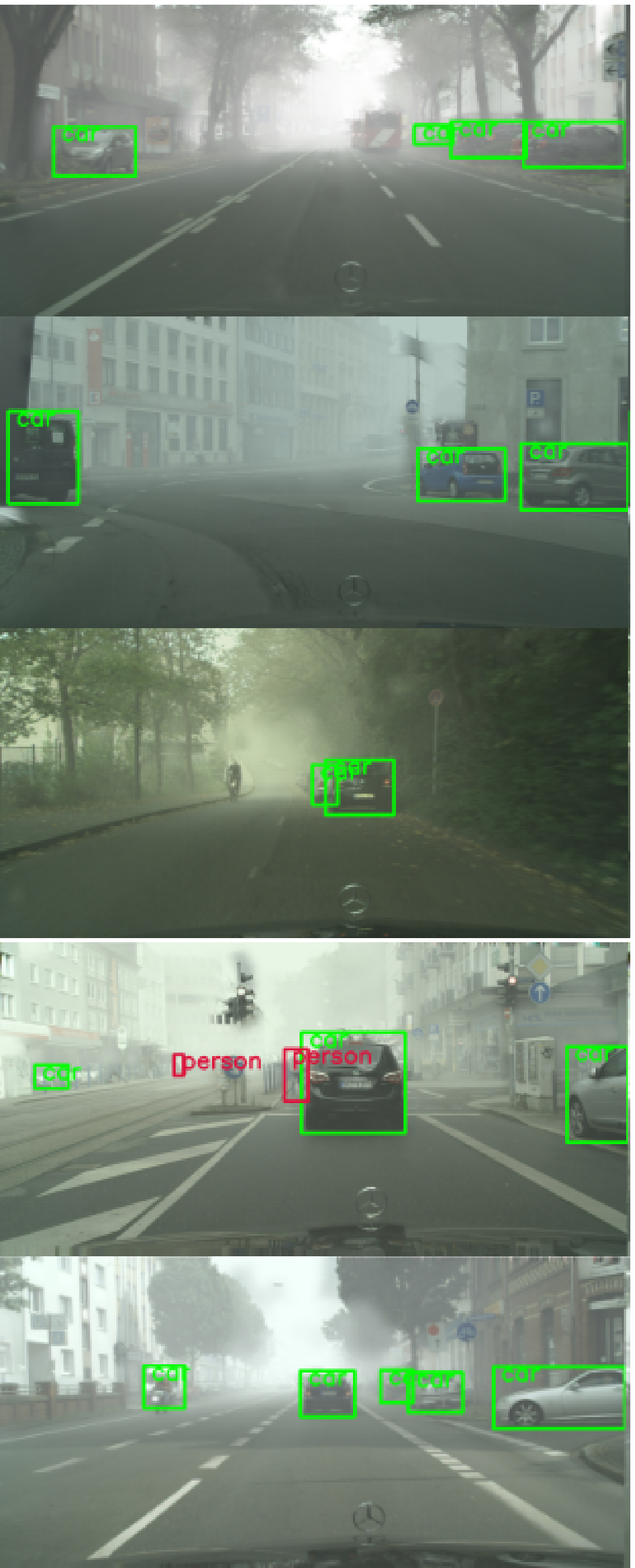}
    \label{cam_fog_swda}
    \end{minipage}}
\hspace{-9pt}
\subfigure[Ours]{
  \begin{minipage}[t]{.198\textwidth}
    \centering
    \includegraphics[width=1\linewidth]{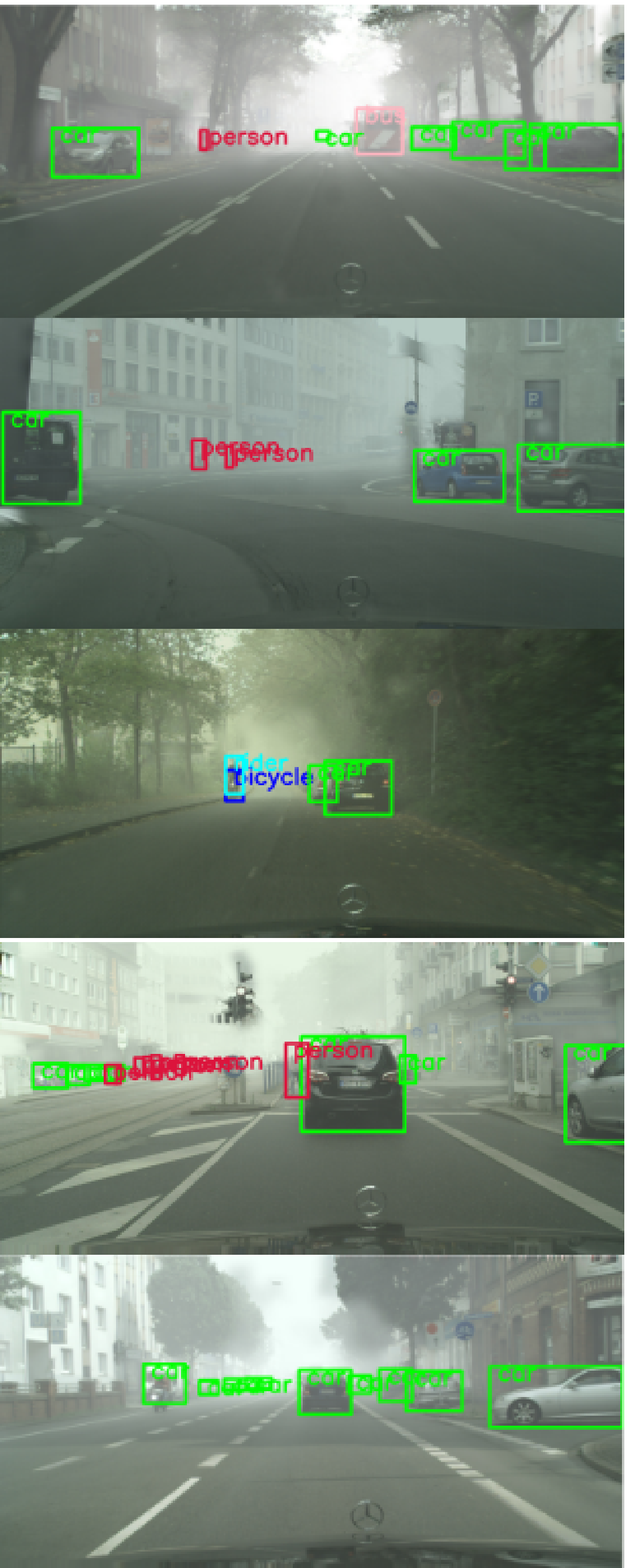}
    \label{cam_fog_ours}
    \end{minipage}}
  \caption{\small Comparison of detection results of different methods on the scenario of Cityscape to Foggy Cityscape.  GT denotes the ground truth bounding boxes.}
  \label{cam_fog}
\end{figure}

\begin{figure}[!htbp]
  \subfigure[GT]{
    \centering
   \includegraphics[width=.198\linewidth]{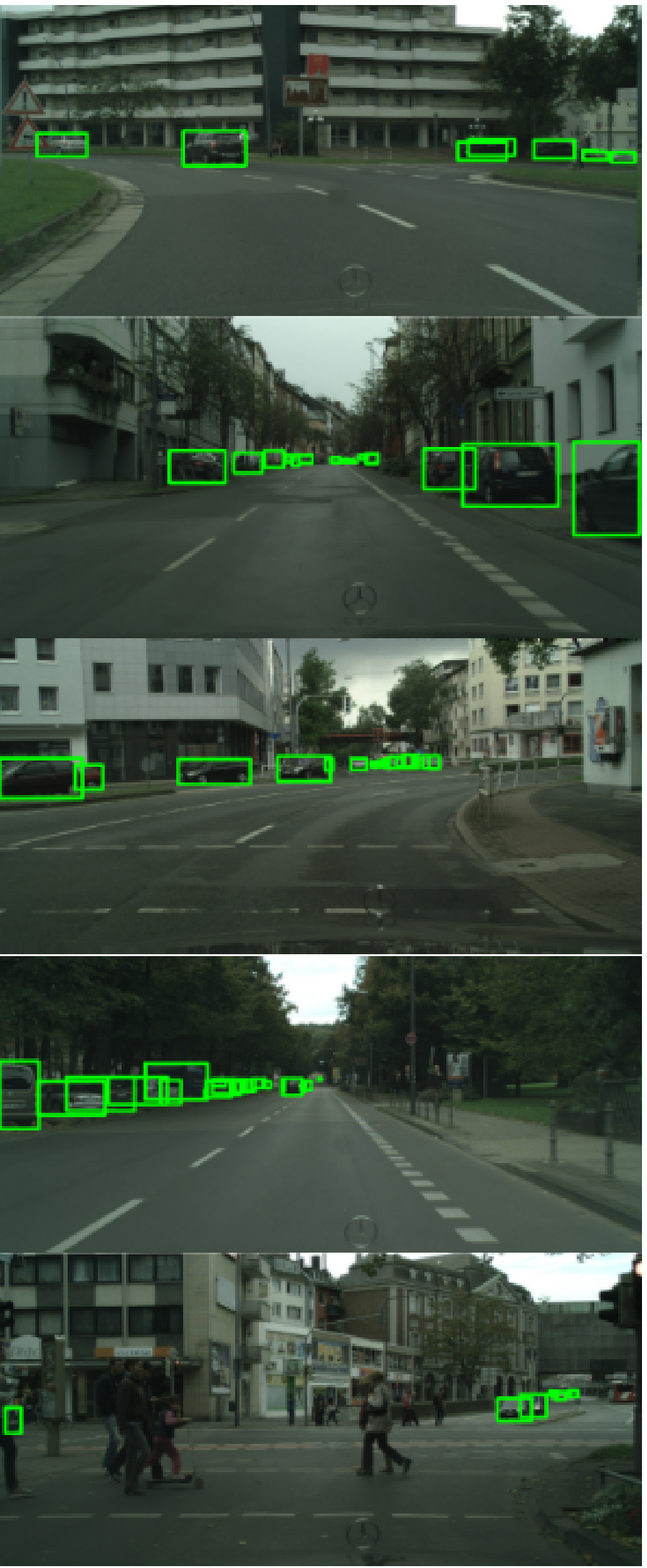}
    \label{cam_car_anno}}
    \hspace{-10pt}
    \subfigure[Source Only]{
    \centering
   \includegraphics[width=.2\linewidth]{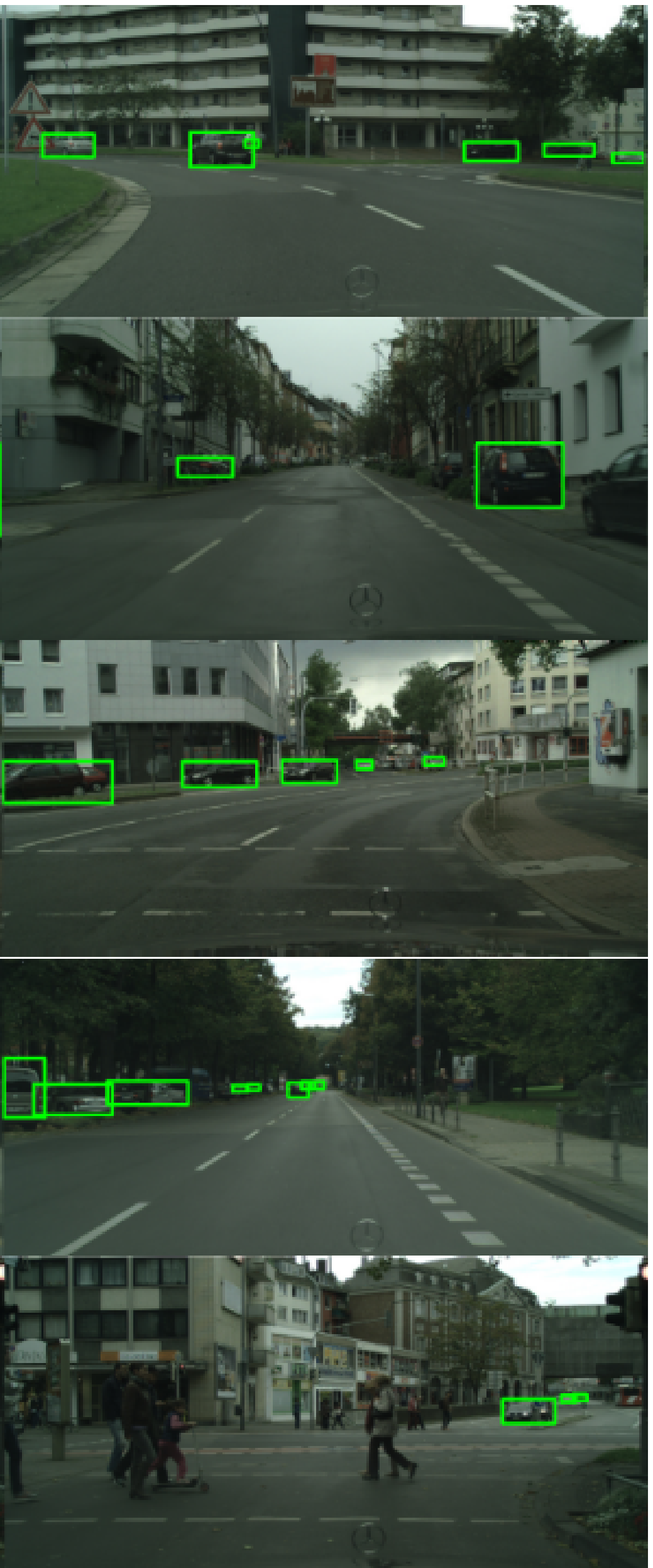}
    \label{cam_car_source_only}}
    \hspace{-10pt}
  \subfigure[DA-Faster]{
    \centering
   \includegraphics[width=.2\linewidth]{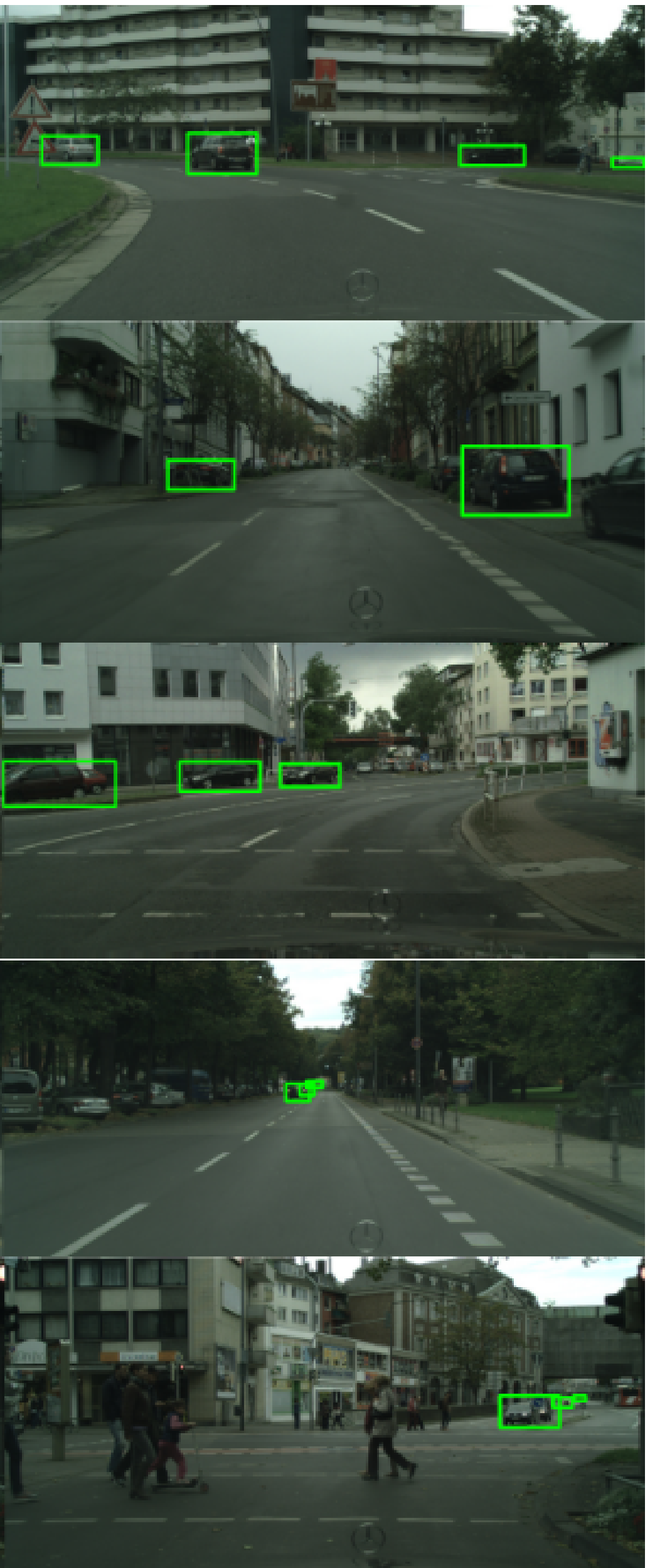}
    \label{cam_car_da_faster}}
    \hspace{-10pt}
  \subfigure[SWDA]{
    \centering
    \includegraphics[width=.2\linewidth]{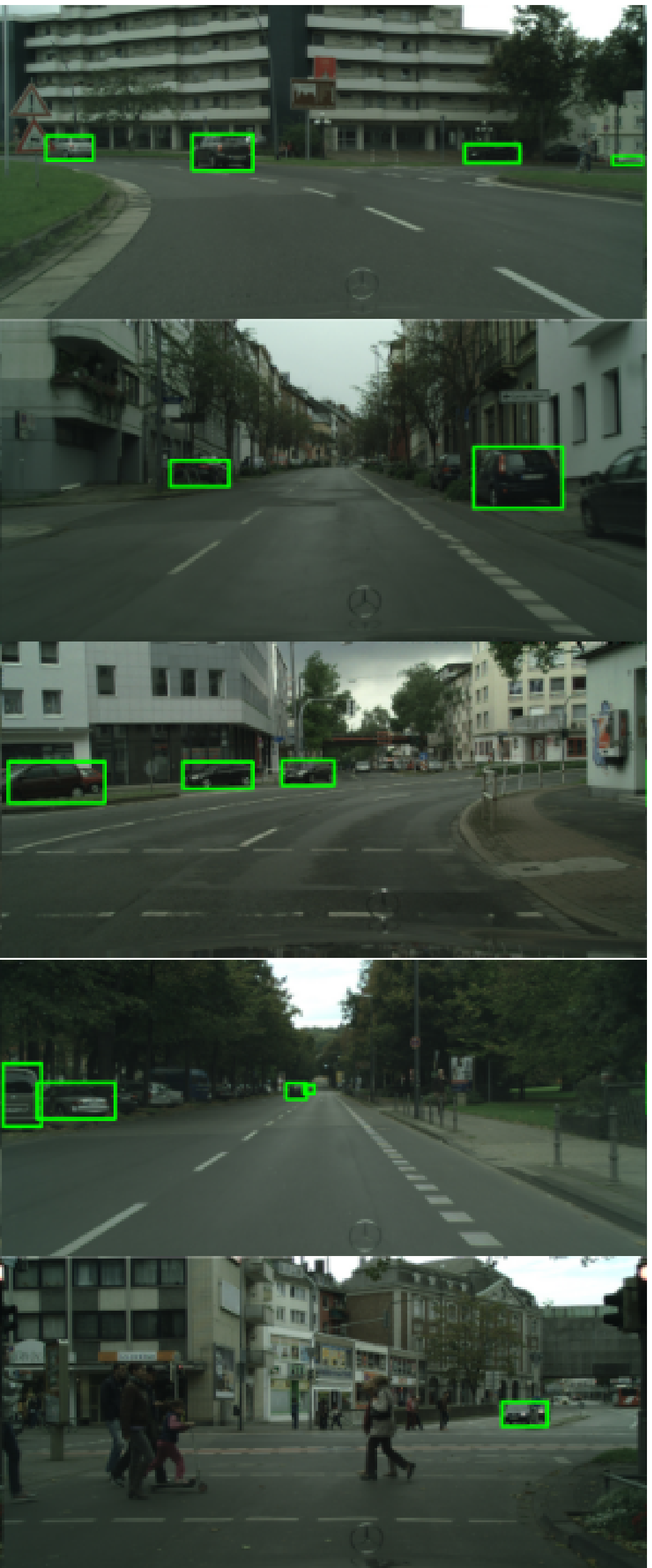}
    \label{cam_car_swda}}
    \hspace{-10pt}
    \subfigure[Ours]{
    \centering
    \includegraphics[width=.2\linewidth]{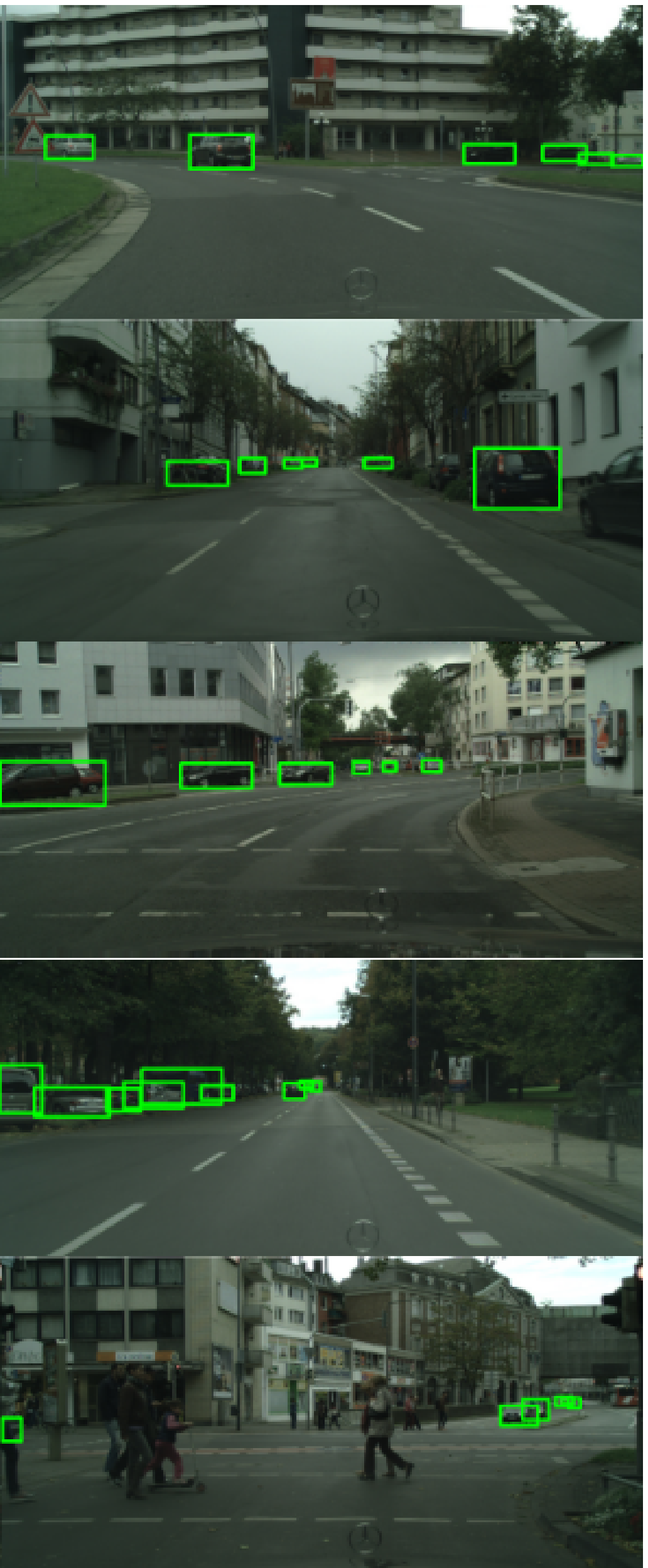}
    \label{cam_car_ours}}
  \caption{\small Comparison of detection results of different methods on the scenario of Sim10k to Cityscape.}
  \label{cam_car}
\end{figure}

\begin{figure}[!htbp]
\centering
\subfigure[GT]{
\centering
\includegraphics[width=.2\linewidth]{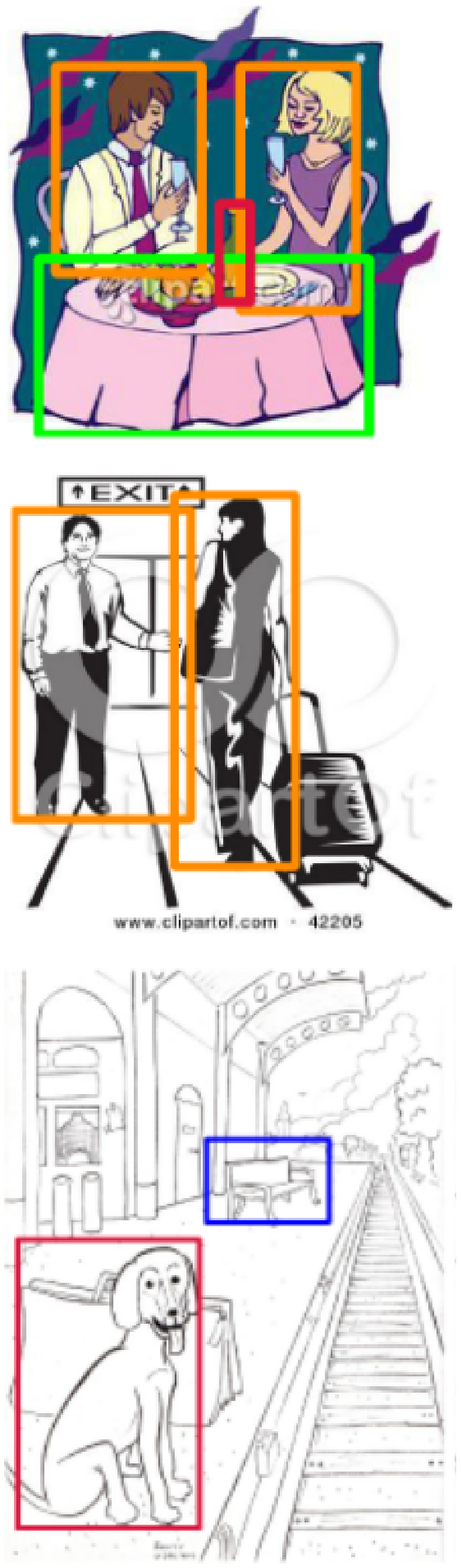}
\label{cam_clip_anno}}
\hspace{-10pt}
\subfigure[Source Only]{
\centering
\includegraphics[width=.198\linewidth]{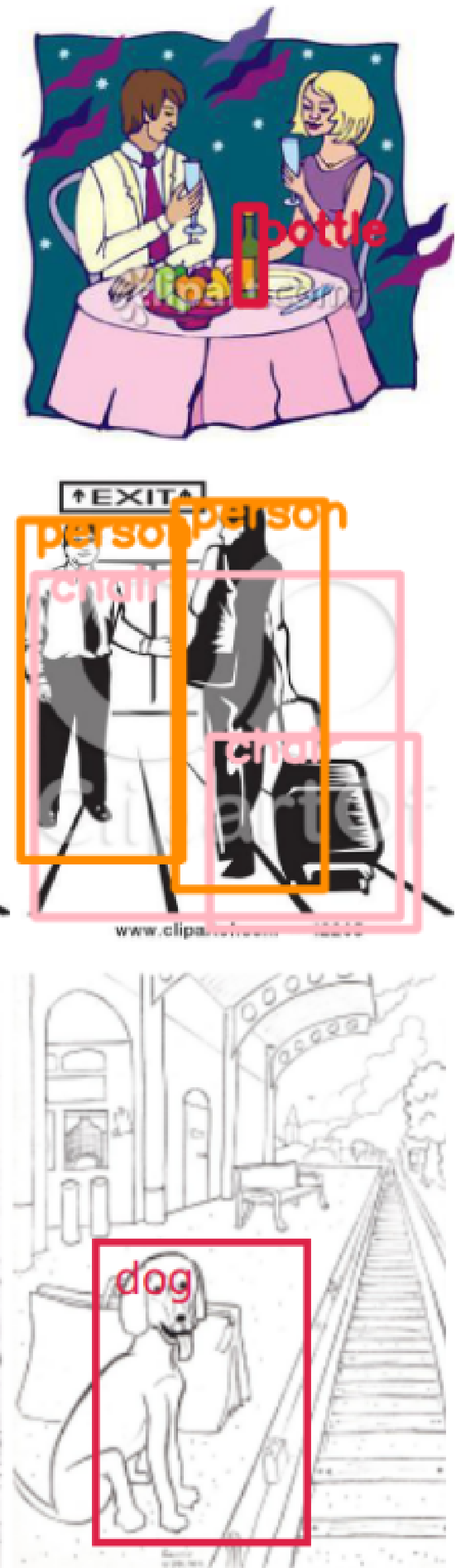}
\label{cam_clip_source_only}}
\hspace{-10pt}
\subfigure[DA-Faster]{
\centering
\includegraphics[width=.2\linewidth]{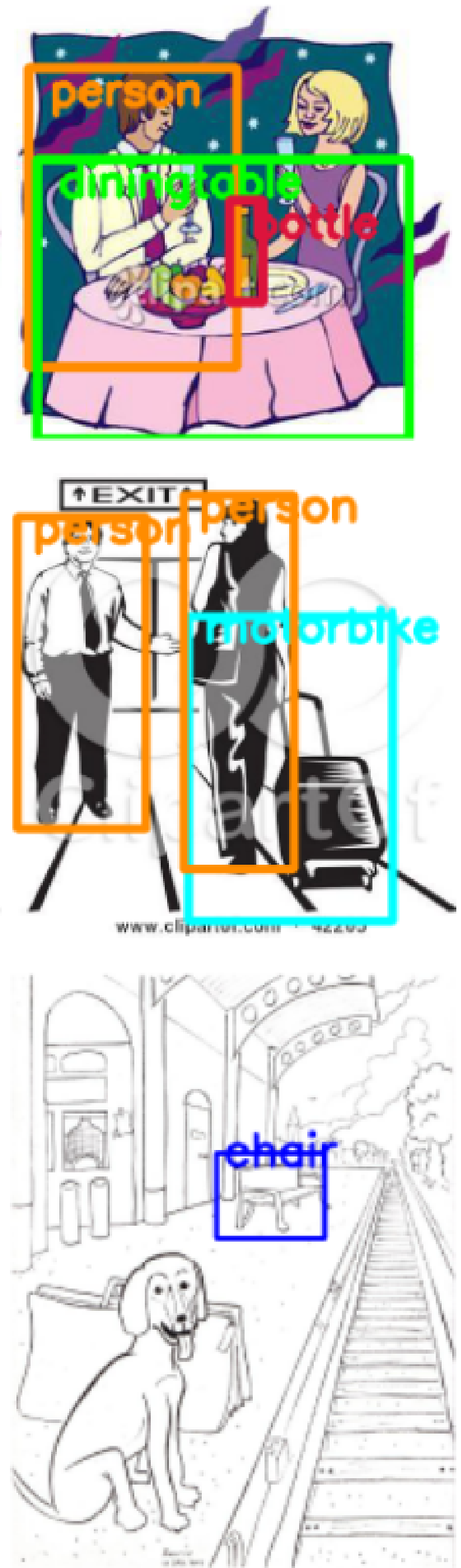}
\label{cam_clip_da_faster}}
\hspace{-10pt}
\subfigure[SWDA]{
\centering
\includegraphics[width=3.49cm]{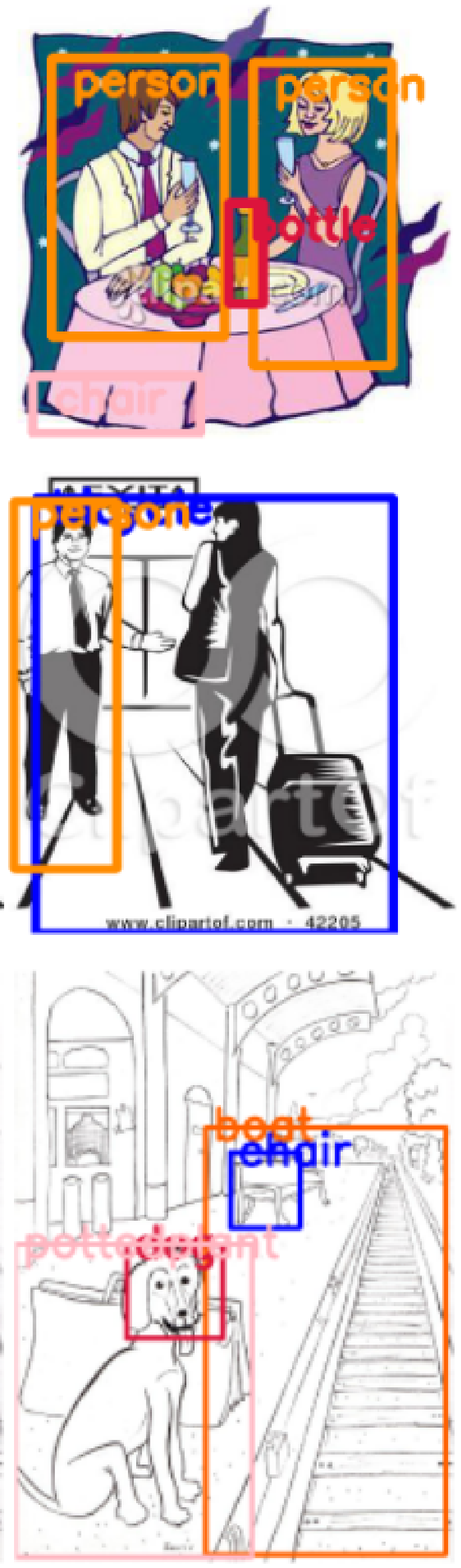}
\label{cam_clip_swda}}
\hspace{-10pt}    
\subfigure[Ours]{
\centering
\includegraphics[width=.194\linewidth]{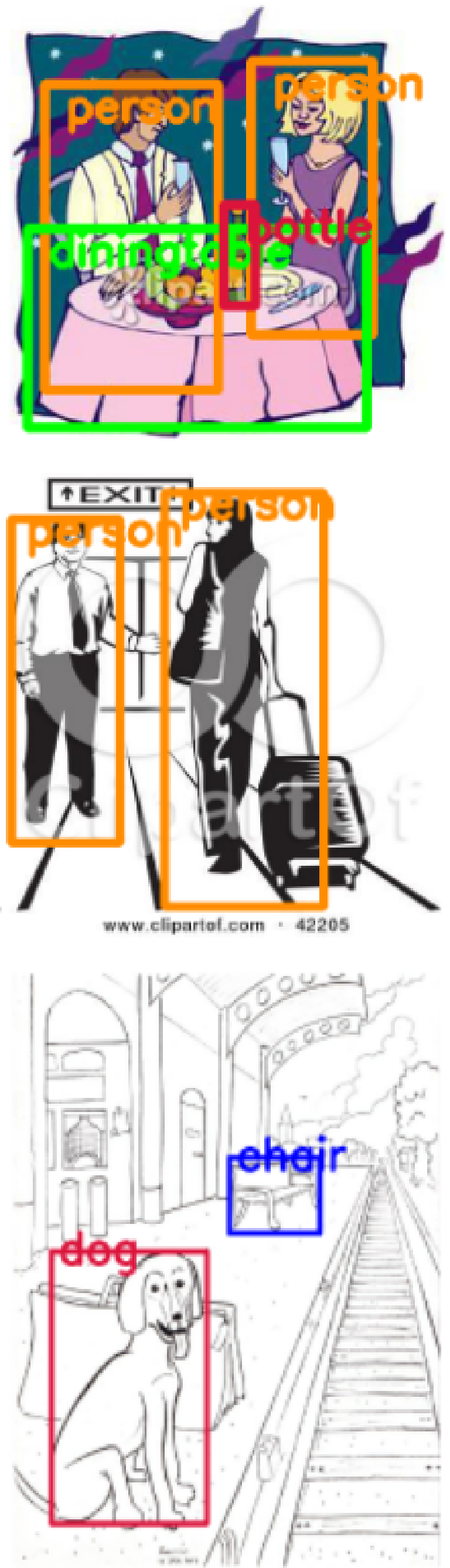}
\label{cam_clip_ours}}
\caption{\small Comparison of detection results of different methods on the scenario of Pascal VOC to Clipart.}
\label{cam_clip}
\end{figure}

\begin{figure}[!htbp]
\subfigure[GT]{
\centering
\includegraphics[width=.189\linewidth]{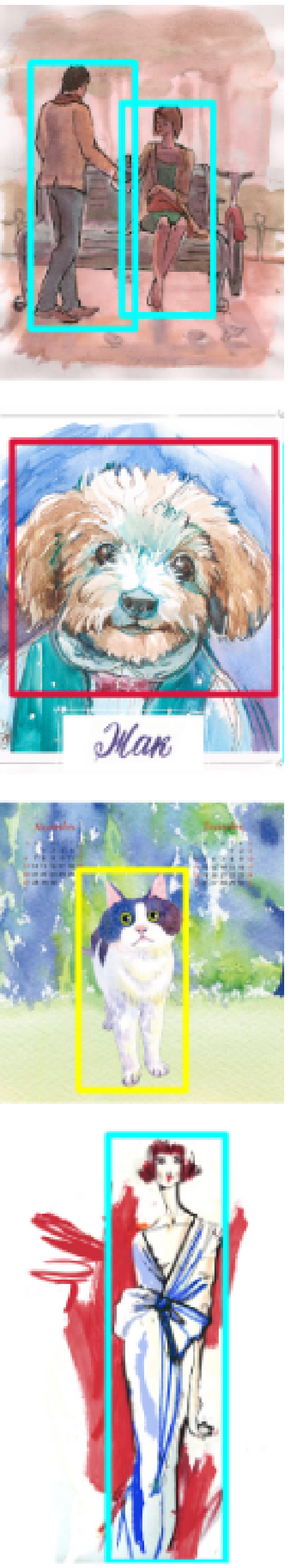}
\label{cam_water_anno}}
\hspace{-9pt}
\subfigure[Source Only]{
\centering
\includegraphics[width=.2015\linewidth]{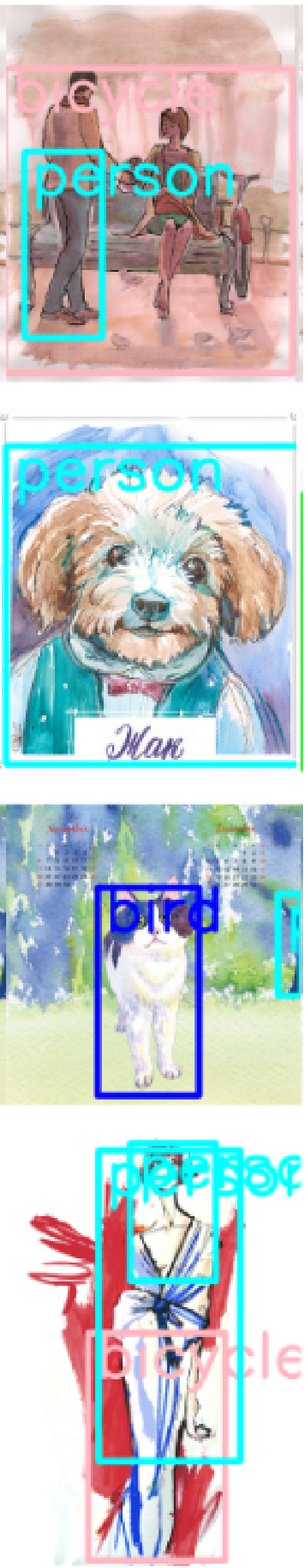}
\label{cam_water_source_only}}
\hspace{-9pt}
\subfigure[DA-Faster]{
\centering
\includegraphics[width=.2055\linewidth]{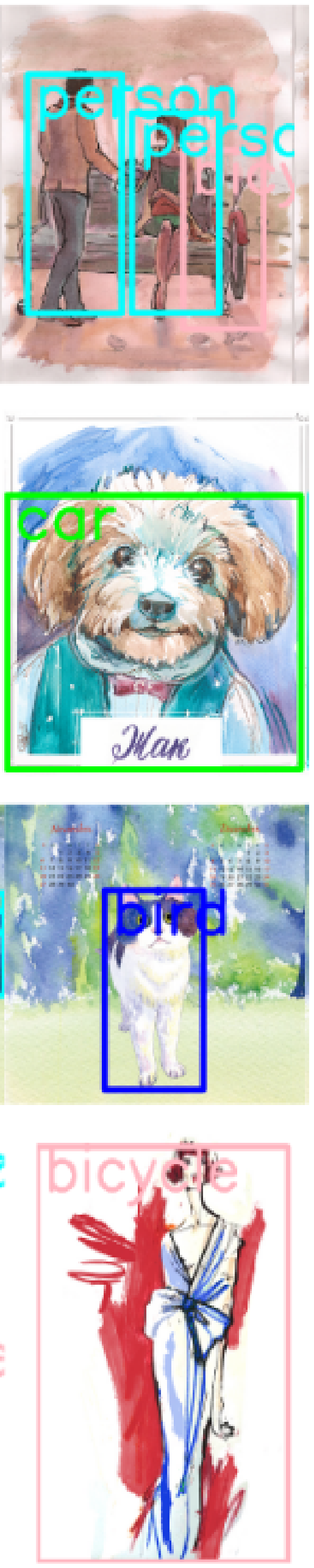}
\label{cam_water_da_faster}}
\hspace{-9pt}
\subfigure[SWDA]{
\centering
\includegraphics[width=.1945\linewidth]{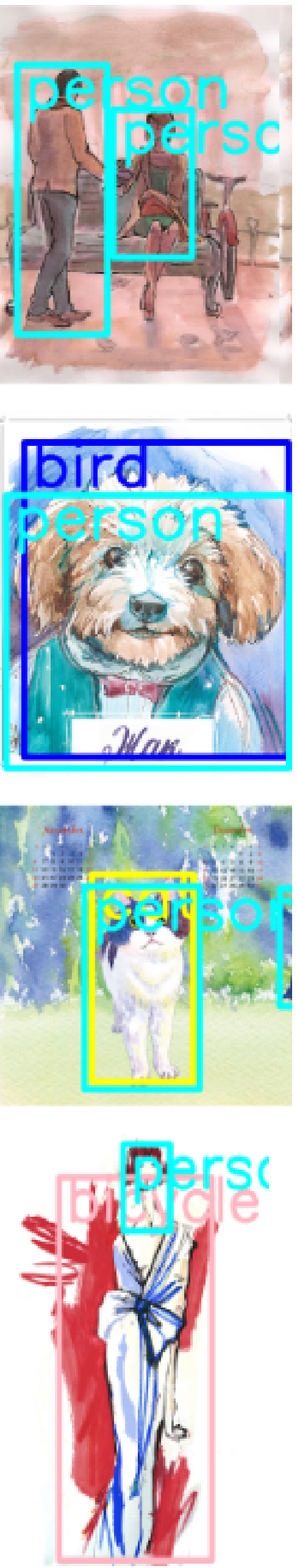}
\label{cam_water_swda}}
\hspace{-9pt}
\subfigure[Ours]{
\centering
\includegraphics[width=.2\linewidth]{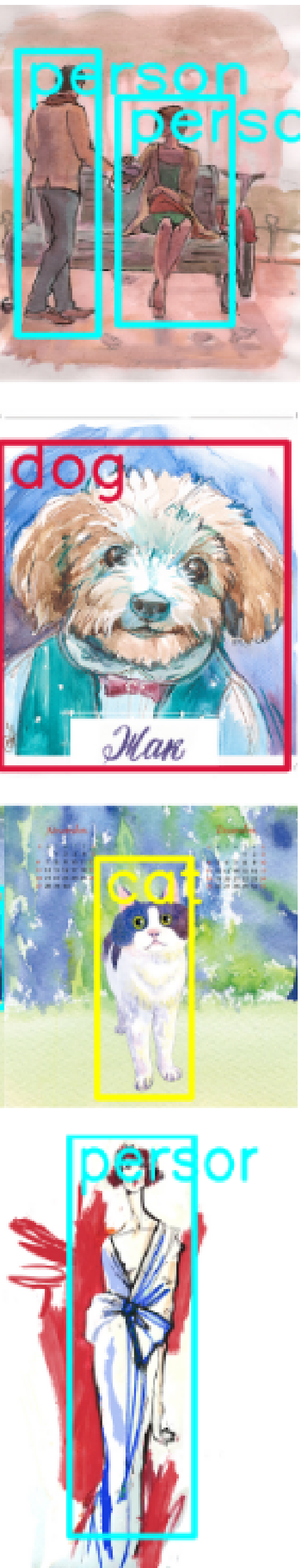}
\label{cam_water_ours}}
\caption{\small Comparison of detection results of different methods on the scenario of Pascal VOC to Watercolor.}
\label{cam_water}
\end{figure}

Comparisons of the detection results of different models on four datasets are depicted respectively in Fig.~\ref{cam_fog},~\ref{cam_car},~\ref{cam_clip},~\ref{cam_water}. Compared to previous methods, our DAA successfully boosts the domain adaptation task, and we observe that better detection performance is achieved in the target domain.

\section{Structure of Discriminators}
\begin{figure}[!htbp]
\centering
  \subfigure[Local Adaptor]{
   \includegraphics[width=.22\linewidth]{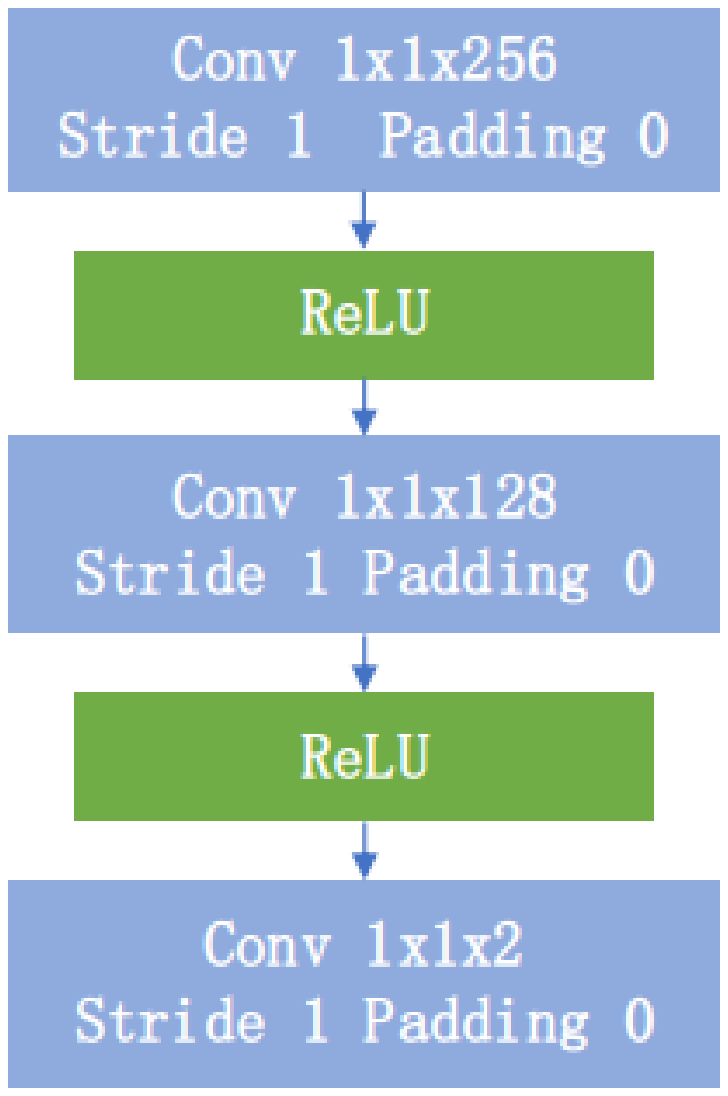}
    \label{adaptor_local}}
  \subfigure[Transition Adaptor]{
   \includegraphics[width=.22\linewidth]{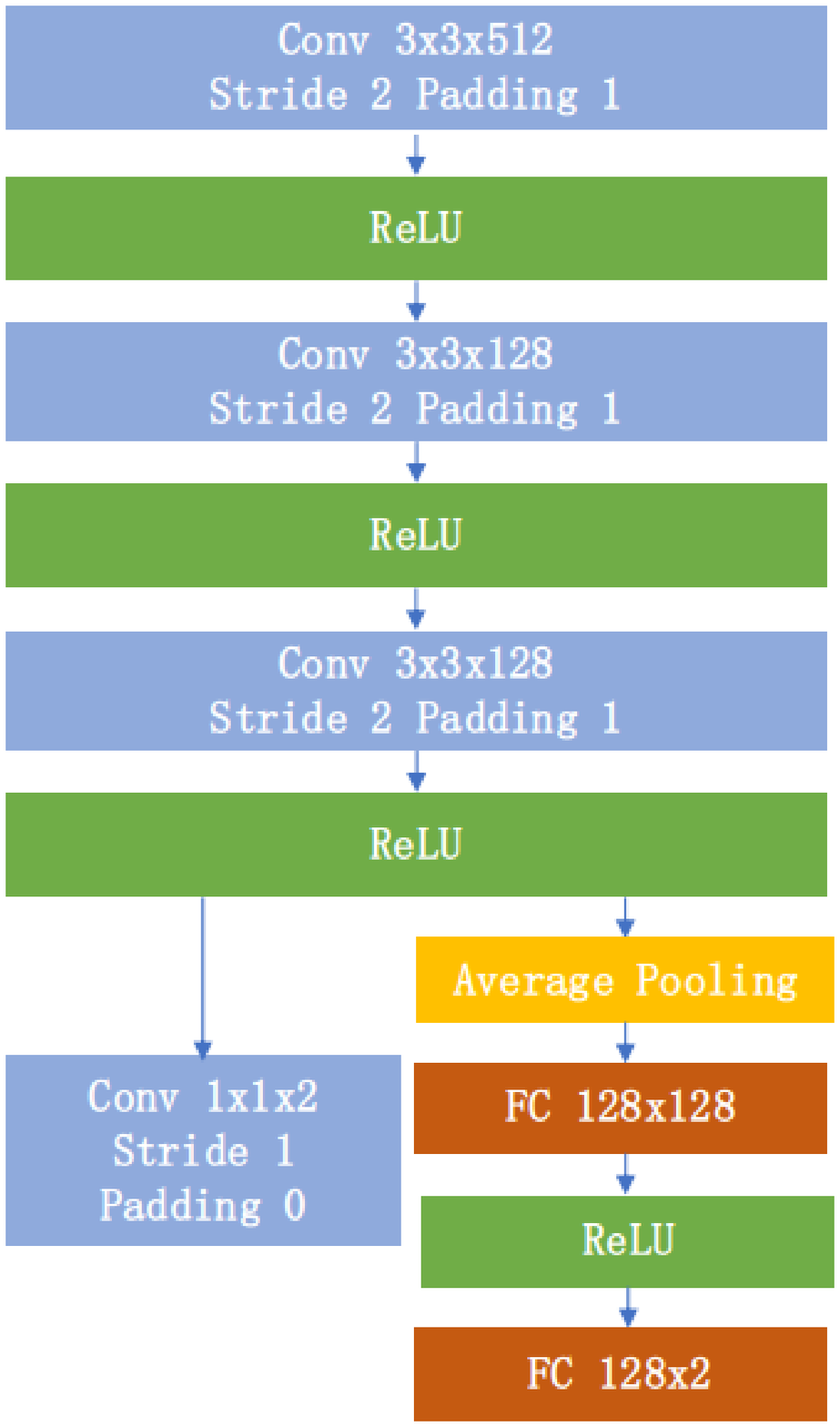}
    \label{adaptor_trans}}
    \subfigure[Global Adaptor]{
   \includegraphics[width=.22\linewidth]{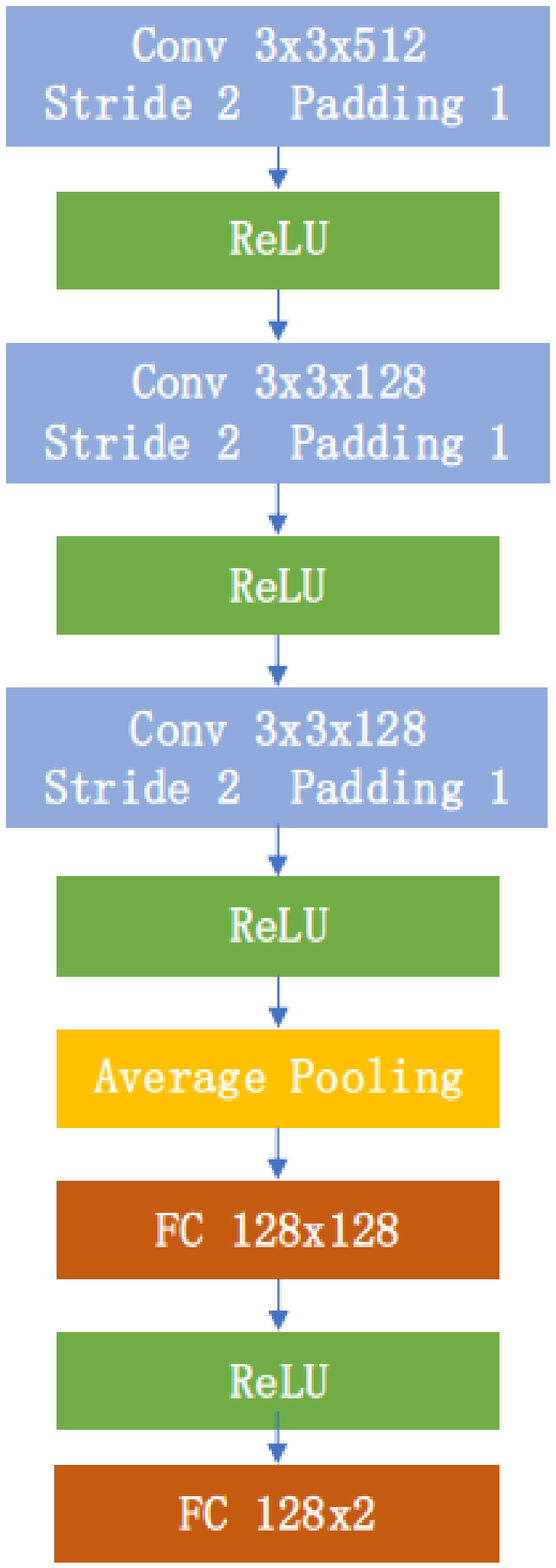}
    \label{adaptor_global}}
  \subfigure[Instance-level Adaptor]{
    \includegraphics[width=.22\linewidth]{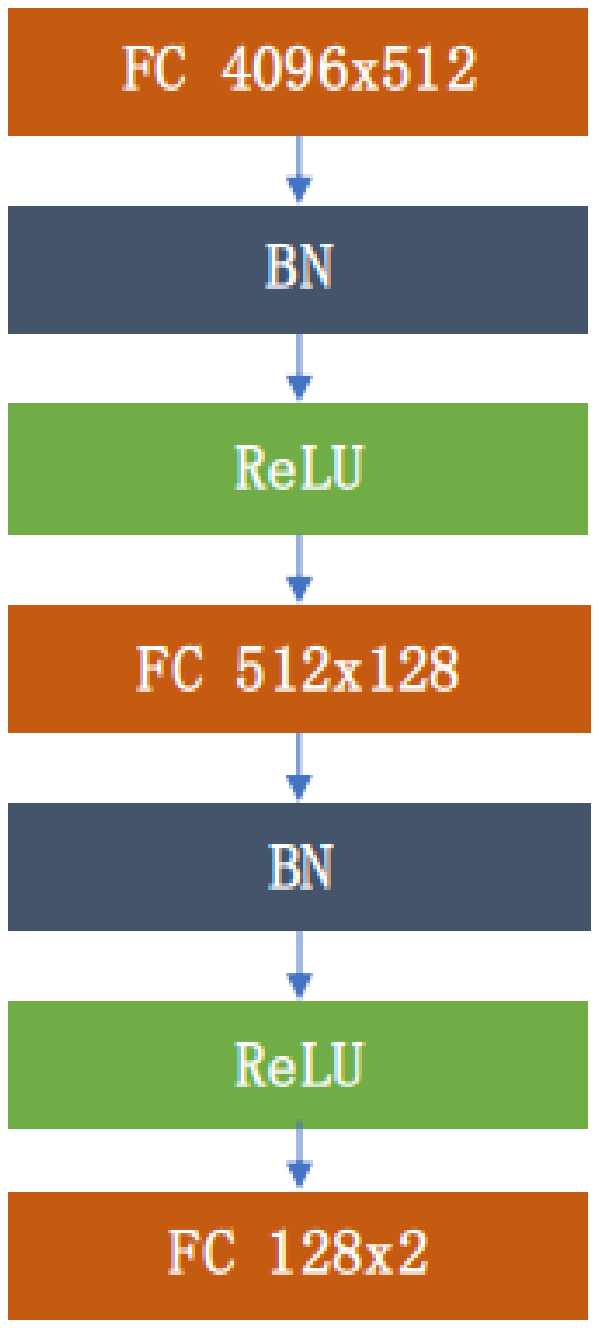}
    \label{adaptor_ins}}
  \caption{\small The Structure of Adaptors}
  \label{adaptors}
\end{figure}

The detailed network structures of our proposed image-level and instance-level domain discriminators are shown in Fig.~\ref{adaptors}. We use the same structure for different backbones of VGG16 and Resnet101, except that the input channels are adjusted to fit the corresponding input feature channels. Here we take VGG16 as an example to illustrate their structures.

\vspace{30pt}

\noindent\textbf{Local Adaptor:} The local adaptor is composed of three $1 \times 1$ convolutional layers with decreasing output channels. After layers 1 and 2, the ReLU is deployed as the activation function. The output of the last convolutional layer will be directly used for calculating the local patch loss.

\noindent\textbf{Global Adaptor:} The global adaptor consists of a convolutional block and a top part. The convolutional block is composed of three $3 \times 3$ convolutional layers with stride 2, while the top part is implemented with two fully-connected layers. An average pooling operation is placed between two parts.

\noindent\textbf{Transition Adaptor:} The global branch of the transition adaptor is the same as the global adaptor. The local branch shares three convolutional layers with the global branch, and a $1 \times 1$ convolutional layer is placed after the shared layers to assign each local patch with a domain prediction.

\noindent \textbf{Instance-level Domain Discriminator:} The instance-level domain discriminator is composed of three fully connected layers followed by batch normalization and ReLU. The object and background discriminators have the identical structure.

\vspace{30pt}

\end{document}